\documentclass[journal]{IEEEtran}
\usepackage{diagbox}
\usepackage{hyperref}
\usepackage[switch]{lineno}
\usepackage{multirow}

\usepackage[flushleft]{threeparttable} 
\usepackage{xcolor} 
\usepackage{amssymb} 
\usepackage{algpseudocode} 
\algnewcommand\algorithmicforeach{\textbf{for each}}
\algdef{S}[FOR]{ForEach}[1]{\algorithmicforeach\ #1\ \algorithmicdo}

\usepackage{lipsum} 
\newcommand{\ignore}[1]{}

\newcommand{\Ind}[1]{\mathbb{I}\left\{{#1}\right\}}
\newcommand{\E}{\mathbb{E}}

\usepackage{graphicx}
\DeclareGraphicsExtensions{.eps,.jpeg,.png}
\usepackage{amsmath}
\ifCLASSOPTIONcompsoc
  \usepackage[caption=false,font=normalsize,labelfont=sf,textfont=sf]{subfig}
\else
  \usepackage[caption=false,font=footnotesize]{subfig}
\fi
\begin{document}
%
\title{Adaptive LiDAR Sampling and Depth Completion using Ensemble Variance}
%
%
%
%

\author{Eyal~Gofer, Shachar~Praisler,~\IEEEmembership{Student Member,~IEEE,}
        and Guy~Gilboa,~\IEEEmembership{Senior~Member,~IEEE}
\thanks{Manuscript received *; revised *; accepted *. Date of publication *; date of current version *. This work was supported in part by *. The associate editor coordinating the review of this manuscript and approving it
for publication was *. E. Gofer and S. Praisler contributed equally to this work. (Corresponding author: E. Gofer.)}%
\thanks{E. Gofer, S. Praisler and G. Gilboa are with the Department of Electrical Engineering, Technion - Israel Institute of Technology, Haifa, Israel, e-mail:
eyal.gofer@ee.technion.ac.il, spraizler@campus.technion.ac.il,
 guy.gilboa@ee.technion.ac.il.}%
\thanks{Digital Object Identifier *}}

\maketitle

\begin{abstract}
This work considers the problem of depth completion, with or without image data, where an algorithm may measure the depth of a prescribed limited number of pixels. 
The algorithmic challenge is to choose pixel positions strategically and dynamically to maximally reduce overall depth estimation error. This setting is realized in daytime or nighttime depth completion for autonomous vehicles with a programmable LiDAR. 

Our method uses an ensemble of predictors to define a sampling probability over pixels.
This probability is proportional to the variance of the predictions of ensemble members, thus highlighting pixels that are difficult to predict. By additionally proceeding in several prediction phases, we effectively reduce redundant sampling of similar pixels.

Our ensemble-based method may be implemented using any depth-completion learning algorithm, such as a state-of-the-art neural network, treated as a black box. In particular, we also present a simple and effective Random Forest-based algorithm, and similarly use its internal ensemble in our design.

We conduct experiments on the KITTI dataset, using the neural network algorithm of Ma et al. and our Random Forest-based learner for implementing our method. The accuracy of both implementations exceeds the state of the art. Compared with a random or grid sampling pattern, our method allows a reduction by a factor of 4--10 in the number of measurements required to attain the same accuracy. 
\end{abstract}


%
\IEEEpeerreviewmaketitle
\section{Introduction}\label{sec:Introduction}

\begin{figure*}[ht]
\centering
\subfloat{%
       \includegraphics[width=7.14in]{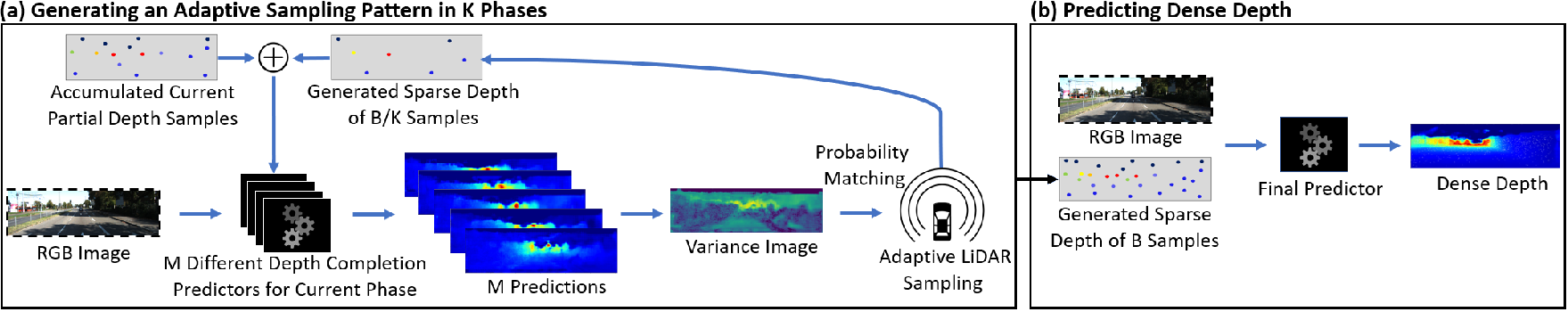}}
\caption{The prediction flow. Given an (optional) RGB image as input, our adaptive depth completion algorithm consists of (a) generating an adaptive sampling pattern of $B$ samples in $K$ phases and (b) predicting dense depth using the RGB input and the generated samples. (a) For each phase, an ensemble of $M$ black-box predictors yields $M$ sets of predictions, which are used to calculate a variance image. A probability proportional to the variance is used to choose the next $B/K$ samples. These samples are added to the existing depth samples as input for the next phase. (b) The generated $B$ samples and the RGB input are used by the final depth completion predictor to produce dense depth.}
\label{fig:general_algorithm}
\end{figure*}

\IEEEPARstart{C}{onstructing} an accurate depth map of a scene is an important computer vision task and an essential technological component in autonomous vehicles. Increasingly, Light Detection And Ranging (LiDAR) is being used to provide a subset of a depth map from which the full map is inferred. In this process, LiDAR measurements may be aggregated with other data such as RGB images, and inference can be performed by either classical image-processing algorithms or machine learning methods. The introduction of LiDAR data has allowed for more accurate depth estimation compared with methods relying on monocular or stereo images alone. In addition, the use of LiDAR enables depth estimation in poor lighting conditions and even in complete darkness.

Common LiDARs operate by scanning a scene periodically along multiple fixed horizontal lines. 
This is done by mechanically rotating the transceivers.
Recently, however, new LiDAR designs are emerging, which allow programmable scanning. They are based on solid-state technologies, where the laser beam is controlled electronically. Thus, instead of following a fixed scanning regime, these new LiDARs could be programmed to measure at dynamically changing points \cite{cheben2018subwavelength,poulton2017coherent}. In particular, scans could be directed at areas of maximal interest, reflecting changing conditions. This technology therefore opens the possibility of obtaining the same quality of depth estimation with far fewer measurements (less acquisition time and power consumption per depth image). Alternatively, given a fixed number of samples, it allows to produce a more accurate reconstructed scene. 
What is needed is a method for choosing the most important points to measure a given scene in an adaptive, data-driven fashion.  
\subsection{A Dynamic Measuring Process}\label{subsec:A Dynamic Measuring Process}
A key observation in the design of the method suggested in this work, and probably of any adaptive depth estimation algorithm, is that some points are more important to estimate correctly than others. Intuitively, important points should include ones that are near the boundary between objects that lie at vastly different distances. A depth map that does not accurately reflect the position of such
boundaries would suffer a large error (e.g., in absolute or RMSE terms).\footnote{Estimating depth at the interior points of objects is of course important too. However, given the reasonable assumption that natural scenes are approximately piecewise linear, as few as three measurements inside each ``piece'' allow for accurate depth estimation through linear interpolation.} 

This notion of importance is static in that it does not depend on the estimation process. Crucially, a statically important point is not necessarily important to \textit{measure}. For example, as more LiDAR measurements near a point are collected, its depth may be estimated accurately without directly measuring it. This example highlights a different notion of importance that is \textit{dynamic}, and reflects the benefit of measuring a point to improving overall depth estimation at a given time.

To complicate matters, one often has image data available, which may be used to estimate boundaries between objects accurately using various computer vision methods. In this case, the importance of LiDAR measurements at boundary points also depends on the available visual cues.

An algorithm for adaptive measurements thus needs to assess the relative dynamic importance of each point given the available information sources. In addition, it should be able to handle the usual challenges of depth completion, namely, complex scenes and noisy data. These considerations motivate an algorithmic solution that is built on top of an existing depth completion algorithm and that defines a notion of dynamic importance based on its performance.

\subsection{Error, Variance, and Ensembles}\label{subsec:Error, Variance, and Ensembles}
Given a depth completion algorithm $A$, consider applying it to a scene and then examining the pixel-wise depth prediction error,
for example, the squared error. Pixels where the error is large are reasonably expected to be those where measurement could help the most.\footnote{Note that given a black-box algorithm, this is a reasonable approximation, but not a guarantee.} Prediction error is thus a natural criterion for static importance. 

Normally, however, depth ground truth is unavailable, meaning that prediction error cannot be used for choosing points to measure. We will need a proxy for the error that is always available. 

Consider then an ensemble of predictors, which are different variants derived from $A$. Given that $A$ is a machine learning algorithm, such variants may be obtained by training $A$ on different subsets or bootstrap samples of the training data.\footnote{Even if $A$ is completely deterministic for a given scene, variants may be created for each scene by taking different samples of the measured points, or by adding random noise to the positions of measured pixels.}
Viewing the depth prediction of a variant at a given pixel as a statistical estimator $P$ of the ground truth $g$, its squared error famously may be expressed as the sum of its (both non-negative) variance and bias. Thus, the variance of $P$ is always a lower bound of the error. Assuming optimistically that $A$ produces predictors with small bias, the variance may serve as a proxy for the error in our suggested error-based measuring scheme.

\subsection{Phased Sampling and Probability Matching}\label{subsec:Phased Sampling}
To use variance as a dynamic notion of importance, we implement two additional mechanisms. First, we apply the ensemble construction in several phases (usually four or eight), where an equal fraction of the measurement budget is used in each phase. Specifically, in each phase an ensemble of predictors is trained given the currently available measurements, and the variance of the predictions of ensemble members is used to select the next pixels to be measured. This results in a hierarchy of ensembles that may then be applied in order to measure pixels in a test image. Phased sampling ensures that we update pixel importance as sampling progresses, to a degree that depends on the number of phases. Choosing the exact number of phases must balance the need for frequent updates against the increased computational cost and the possibility of overfitting.

Second, rather than measure the points with the highest variance in each phase, we define a probability that is proportional to the variance, and use it for sampling, an approach known as \textit{probability matching} (see, e.g., \cite{chapelle2011empirical}). This ensures that if several regions with high-variance points exist, we will tend to measure points from all of them, rather than focus on the ``best'' points, which may be concentrated in a single region, or even a small part of it. 

Once the sampling process is executed on the training set, a final predictor may be trained on the data, which includes all the sampled depths. This final predictor may be applied to a test image that has been similarly augmented with depth measurements. An overall view of the prediction process is given in Figure~\ref{fig:general_algorithm}.

\subsection{Our Contributions}\label{subsec:contributions}
In this paper we present the following main novelties and contributions:
\begin{enumerate}
    \item We propose a new, very general, adaptive sampling method for LiDARs, which leverages principles from statistics and active learning.
    The proposed method can enhance any learning-based depth completion algorithm, where an ensemble can be constructed. We showcase this by applying the same generic sampling method to two completely different completion algorithms: one based on a deep neural network and the other on Random Forest. 
    \item A new depth completion algorithm based on Random Forest is introduced. It is very lean, is based on only 26 hand-crafted features per pixel and needs a very small amount of data for training. In our experiments, its performance with adaptive sampling was on a par or even slightly better than the neural net-based implementation. 
    \item Extensive experiments on the KITTI depth completion data \cite{Uhrig2017THREEDV} show our sampling algorithm outperforms grid and random sampling, as well as state-of-the-art adaptive sampling methods \cite{wolff2020icra,bergman2020deep}. Compared with random sampling or sampling on a grid, our method requires a factor of \hbox{4--10} fewer measurements to achieve the same RMSE, where the exact factor depends on the underlying depth completion algorithm and the target error level. 
    \item Our algorithm works well also for the case where no RGB is available (unguided depth completion), unlike \cite{wolff2020icra} and \cite{bergman2020deep}. Thus LiDAR sampling can be enhanced also at night or when visibility is poor.
\end{enumerate}

\section{Related Work}\label{sec:Related Work}
Adaptive depth sampling for depth completion is naturally considered as a task in computer vision but should also be seen in the wider research context of adaptive sampling and active learning. Both these perspectives are given in this section.
\subsection{Depth Completion}\label{subsec:Depth Completion}
Depth completion involves estimating a dense depth image from a partial one, usually with the aid of an additional RGB image to help overcome the loss of spatial information. Throughout the years, some classical approaches have been used for this task, but most of these works handled inputs of low resolution dense depth \cite{park2010depth, park2014high, lu2015sparse} and almost-dense depth that is highly noisy or missing some data \cite{wang2008stereoscopic, camplani2012efficient, shen2013layer, lu2014depth}. Only few works \cite{drozdov2016robust, ma2016sparse, ku2018defense} have dealt with the more challenging problem of working on a sparse depth, namely, a scattered, small percentage of valid depth pixels.
In recent years, with the wider use of deep learning methods, dealing with sparse depth has become more common. These frameworks have become dominant, showing state-of-the-art results.

Many authors have proposed new CNN architectures, mainly variations of encoder-decoder, to learn directly how to complete depth, based on local and global connections between the depth pixels, and if such exist, also between the RGB image and the depth pixels. Some works \cite{ma2018sparse, chen2018estimating, cheng2018depth, wang2018multi, shivakumar2019dfusenet} focused on finding better variants of CNNs, while others created designated modules for them. Such modules can be found, for example, in the works of Huang et al. \cite{huang2019hms}, who introduced sparsity-invariant operations for handling sparse inputs and sparse feature maps, and Chen et al. \cite{chen2019learning}, who created 2D-3D fuse blocks for better extraction of joint features. Tang et al. \cite{tang2019learning} developed a guided-convolution module that generates content-dependent and spatially-variant kernels, while only recently, Lee et al. \cite{lee2020deep} introduced a new cross guidance module to help share information between the RGB and the sparse depth.

Other researchers suggested a multi-task approach to benefit from the relations between the tasks (often, as constraints). See, e.g., Lee et al. \cite{lee2019depth} who additionally computed surface normals, Eldesokey et al. \cite{eldesokey2019confidence} and Van et al. \cite{van2019sparse} who applied a confidence map to refine the depth result, and Qiu et al. \cite{qiu2019deeplidar} and Xu et al. \cite{xu2019depth} who used both.

It is worth mentioning that most networks were trained in a supervised manner, but some self-supervision can also be found \cite{ma2019self, qu2020depth}.

\subsection{Adaptive Depth Sampling}\label{subsec:Depth Sampling}
Various sampling techniques have been intensely explored over the years for tasks such as scene reconstruction, noise reduction and compact representations \cite{eldar1997farthest, zhu2015adaptive, dovrat2019learning, dai2019adaptive}. These works involved sampling from fully-available data rather than querying for missing data, namely, a different setting than ours.
For the problem of depth sampling, despite developments in depth completion methods, only few works have tried sampling patterns other than random, LiDAR scans (horizontal rows) or grid, which are all non-adaptive.

Early guided depth sampling has been introduced for disparity map reconstruction. These works focused on sample areas with high magnitudes of depth gradients, but did not deal with very low (below 5\%) sampling budgets. For example, Hawe et al. \cite{hawe2011dense} assumed that disparity discontinuities coincide with image intensity edges, and therefore applied an edge detector to the image and divided the budget between the edgy areas and the smoother parts. Liu et al. \cite{liu2015depth} suggested obtaining an estimation of the disparity using half of the budget (with a uniformly random pattern) and then improving the initial disparity estimation by sampling the other half of the budget along the depth gradients.

Recently, adaptive sampling algorithms for depth completion were introduced following the emergence of new optical machinery \cite{sun2013large, tasneem2018directionally} that allowed sampling irregular patterns more precisely. Wolff et al. \cite{wolff2020icra} proposed an image-driven sampling and reconstruction strategy based on dividing the image into approximately piecewise segments (using super-pixels), followed by sampling each center of mass and filling the entire segment with the sampled depth. This method required one-fourth to one-third of the samples for a given reconstruction RMSE relative to random or grid patterns. Another work by Bergman et al. \cite{bergman2020deep} introduced deep neural network for end-to-end sampling and reconstruction. A grid pattern was taken as a prior, and then an importance vector flow field was used to move the initial location of the samples into the final, more interesting areas.

\subsection{Active Learning and Generic Adaptive Sampling}\label{subsec:Active Learning}
Active learning (AL) allows an algorithm to query for labels, in contrast to the more common form of supervised learning. Active learning and adaptive sampling have been the target of extensive research in the machine learning and the design of experiments literature \cite{hannekeActiveSurvey, settles2009active,liu2018survey}. Both heuristic and theoretical results have focused more on classification than on regression. 

In computer vision, AL has also been applied primarily for classification, with the purpose of reducing the human effort of annotating images for segmentation and object detection. See, e.g., \cite{vandoni2019evidential,polewski2015active}, and \cite{konyushkova2019geometry}, which also mentions additional examples. Such applications can also be found in the context of autonomous driving \cite{gu2017embedded,jiang2018integrating,feng2019deep}. Of special interest is the work of Feng et al. \cite{feng2019deep}, which used AL to select data for annotation as part of training an object detector using RGB and LiDAR data. The authors applied, among other methods, ensembles of deep networks for classification, and used the difference in their predictions as the basis for several measures of uncertainty. They observed improvement over random sampling not only in classification, but also, indirectly, in the MSE of localization. In a recent thesis of Rai \cite{rai2019monocular}, AL was used in the context of monocular depth estimation. Emphasis in the training was given to small regions with distinct RGB features, of high probability to be near depth discontinuities.

For regression, variance reduction has been suggested as a goal for AL by Cohn et al. \cite{cohn1996active}, being a component of the error. Cohn \cite{cohn1997minimizing} also examined minimizing the bias, and noted that ideally one would minimize the squared error.

The Query by Committee (QBC) algorithm \cite{seung1992query,freund1997selective} uses an ensemble of predictors to decide probabilistically which sample to query for a label at each time (see also \cite{AbeM98}); the higher the disagreement between the predictors, the higher the probability. While this algorithm was defined for classification, it is natural to extend it to regression and use the variance of ensemble element predictions on a sample as a measure of disagreement \cite{burbidge2007active}, as we do in this work. Variance-based sampling was used successfully by Borisov et al. \cite{borisov2011active} for classification, where it was suggested for regression as well. The work of Douak et al. \cite{douak2012active} showed that active learning based on ensemble variance has a better learning curve than random sampling in a spectroscopy data regression task.

Many authors have noted that considering disagreement alone may result in sampling that is not diverse and representative enough. Increased diversity is desirable for preventing a difficult region from being repeatedly sampled even after it is already ``understood''. Unrepresentative points may simply be outliers. Some works gave methods to account for diversity and representativeness \cite{borisov2011active,douak2012active,kee2018query,wu2018pool,park2019active}. Of those works dealing with regression, that of Douak et al. \cite{douak2012active} showed no improvement over their variance-only approach. In the algorithm of Wu \cite{wu2018pool}, k-means clustering was used on top of QBC to add diversity and representativeness. While showing improved results, this algorithm seems infeasible for large datasets. The method of Park and Kim \cite{park2019active} also used clustering-based diversity and representativeness quantities which it added to a Laplacian regularized least squares optimization objective. While showing some improvement over QBC, this method also has a scalability problem. We comment that performing several iterations of sampling and retraining, as we do, serves to reduce the error (which for MSE includes the variance) of points in regions that are already represented in the measured set, implicitly obtaining diversity.

An alternative approach to variance reduction is to find which sample could minimize the generalization error if added to the training set. Such an approach is typically too expensive \cite{settles2009active}, although approximations are feasible for restricted cases in binary classification (e.g., \cite{roy2001toward,polewski2015active}). More recently, Konyushkova et al. \cite{konyushkova2017learning} presented a method for learning error reduction for a given classifier.

The work of K{\"a}ding et al. \cite{kading2018active} provided an AL method for regression that finds the sample maximizing change in the model output, which they can calculate when using a gaussian process regressor. They showed a consistent advantage over other AL methods, including variance reduction, on image data. They comment on the strength of passive sampling, which comes second only to their method. We note that Wu \cite{wu2018pool} found random sampling inferior to all AL methods tested there, and our methods beat random sampling as well.

The approach of sampling the point with the maximal predicted error was considered in the design of experiments literature \cite{aute2013cross}. These authors trained a kriging model to predict the absolute error on all points, and picked the one with maximal absolute error, conditioning that it be far enough from all points already labeled (diversity).

We conclude by noting that the theory of AL for regression is still not well-developed. A negative result \cite{willett2006faster} showed that even for the family of Lipschitz regression functions with added i.i.d. gaussian noise, AL cannot improve the minimax rate of non-adaptive sampling by more than a constant factor. The same authors showed improvements for more restrictive cases. More recently, Goetz et al. \cite{goetz2018active} showed the optimality of a sampling method for a class of learners (Mondrian trees). Their algorithm starts with a random sample and continues based on the variance in the leaves of the tree. We point out that the above negative result does not preclude the utility of AL in concrete settings or on average.

\section{Methods}\label{sec:Methods}
A depth completion algorithm is tasked with reproducing the true depth of each pixel in a scene given a partial set of (possibly noisy) depth measurements. The algorithm usually has color images as input as well (the so-called RGBd scenario), but may have only depth information (the d scenario). 

For training purposes the algorithm is given a collection of $n$ scenes $\mathcal{I}=\{(I_i,G_i,D_i)\}_{i=1}^n$, where $I_i$ maps pixels to color, $G_i$ maps them to true depth, and $D_i$ to noisy depth measurements for some subset of the pixels, and otherwise to $-1$. For prediction, $G_i$ is either unknown or usable only for evaluation, and the algorithm outputs an estimate $\widehat{G}_i$ of the ground truth for every $i$. We note that the ground truth itself may be available only for some of the pixels, as it is for the KITTI dataset, in which case measurement is further restricted to this available set, as is evaluation.

In an \textit{adaptive} depth completion scenario, which we consider here, each $D_i$ starts as a trivial map, and is updated after each measurement choice the algorithm makes for scene $i$. The algorithm is allowed a total \textit{budget} of $B$ measurements per scene. 

We propose a probabilistic algorithmic solution for the adaptive depth completion problem, based on ensemble variance. A generic construction will be given along with two concrete implementations.

\subsection{A Generic Algorithm}\label{subsec:generic algorithm}

The general training framework of our algorithm comprises several elements. First, an ensemble of predictors is created using a given depth completion algorithm. These predictors are trained on different subsets or bootstrap samples of the training set, resulting in different variants of the same predictor. Second, each ensemble member yields predictions for each training image pixel and the variance of predictions per pixel is computed. Third, the pixels to measure are sampled for each scene with probability proportional to the variance. 

This process is repeated several times, and for each repetition, or \textit{phase}, an equal fraction of the measurement budget is used. We comment that the sampling process does not allow repetitions. Once all measurements are made, a final depth completion predictor is trained on the whole training set, possibly, but not necessarily, using the same algorithm employed for the ensembles.

For testing on new images, the trained ensembles are similarly used to select points to measure, and the final predictor is used for depth completion. The detailed generic procedure is given in Figure~\ref{fig:generic algorithm}. Note that in the depth-only scenario, the first phase of sampling practically picks points uniformly at random.

For both training and testing, the phases are inherently sequential, since each phase depends on information produced in the previous phase. Within each phase, however, computations for each ensemble member are independent of the rest of the ensemble, meaning that they can be parallelized, if so desired.

The algorithm trains a final predictor and a total of $MK$ ensemble members, where $M$ is the size of each ensemble and $K$ is the number of phases. The total time needed to train or apply the ensemble members is a multiple of the time needed for a single member, which itself depends on the black box used. This multiple is $MK$ if the computation is fully sequential, or $K$ if parallelization is applied.

\begin{figure}[ht]
\begin{algorithmic}[1]
\Procedure {PhasedVarPM-Train}{}
\State \textbf{Input:} depth completion algorithms $A$, $A_f$, 
\State scene set $\mathcal{I}=\{(I_i,G_i,D_i)\}_i$, budget $B$, 
\State number of phases $K$, ensemble size $M$
\For {$m \leftarrow 1, M$}
\State $\mathcal{I}_m \leftarrow$ $m$-th set of images selected from $\mathcal{I}$
\EndFor
\For {$k \leftarrow 1, K$}
\For {$m \leftarrow 1, M$}
\State $\mathcal{P}_{k,m} \leftarrow$ train predictor using $A$ and $\mathcal{I}_m$
\EndFor
\For {$i \leftarrow 1,|\mathcal{I}|$}
\ForEach {pixel $x$ of $I_i$}
\State apply $\mathcal{P}_{k,1},\ldots,\mathcal{P}_{k,M}$ to $x$
\State $v_i(x) \leftarrow$ variance of the predictions 
\EndFor
\State $\pi_i(x) \leftarrow$ probability proportional to $v_i(x)$
\State sample $B/K$ new pixels from $\pi_i$, update $D_i$ 
\EndFor
\EndFor
\State $\mathcal{P}_f \leftarrow$ train final predictor using $A_f$ and $\mathcal{I}$
\State \textbf{return} $\mathcal{P}_f$ and $\{\mathcal{P}_{k,m}\}$ for every $k$ and $m$
\EndProcedure
\Statex
\Procedure {PhasedVarPM-Test}{}
\State \textbf{Input:} scene set $\mathcal{I}=\{(I_i,G_i,D_i)\}_i$, predictors $\mathcal{P}_f$, \State $\{\mathcal{P}_{k,m}\}$, budget $B$
\For {$k \leftarrow 1, K$}
\For {$i \leftarrow 1,|\mathcal{I}|$}
\ForEach {pixel $x$ of $I_i$}
\State apply $\mathcal{P}_{k,1},\ldots,\mathcal{P}_{k,M}$ to $x$
\State $v_i(x) \leftarrow$ variance of predictions 
\EndFor
\State $\pi_i(x) \leftarrow$ probability proportional to $v_i(x)$
\State sample $B/K$ new pixels from $\pi_i$, update $D_i$ 
\EndFor
\EndFor
\State $\{\widehat{G}_i\}_i \leftarrow$ apply $\mathcal{P}_f$ to $\mathcal{I}$
\State \textbf{return} $\{\widehat{G}_i\}_i$
\EndProcedure
\end{algorithmic}
\caption{A generic algorithm for adaptive depth completion. %
}
\label{fig:generic algorithm}
\end{figure}

\begin{figure*}[t]
\centering
\includegraphics[width=7.14in]{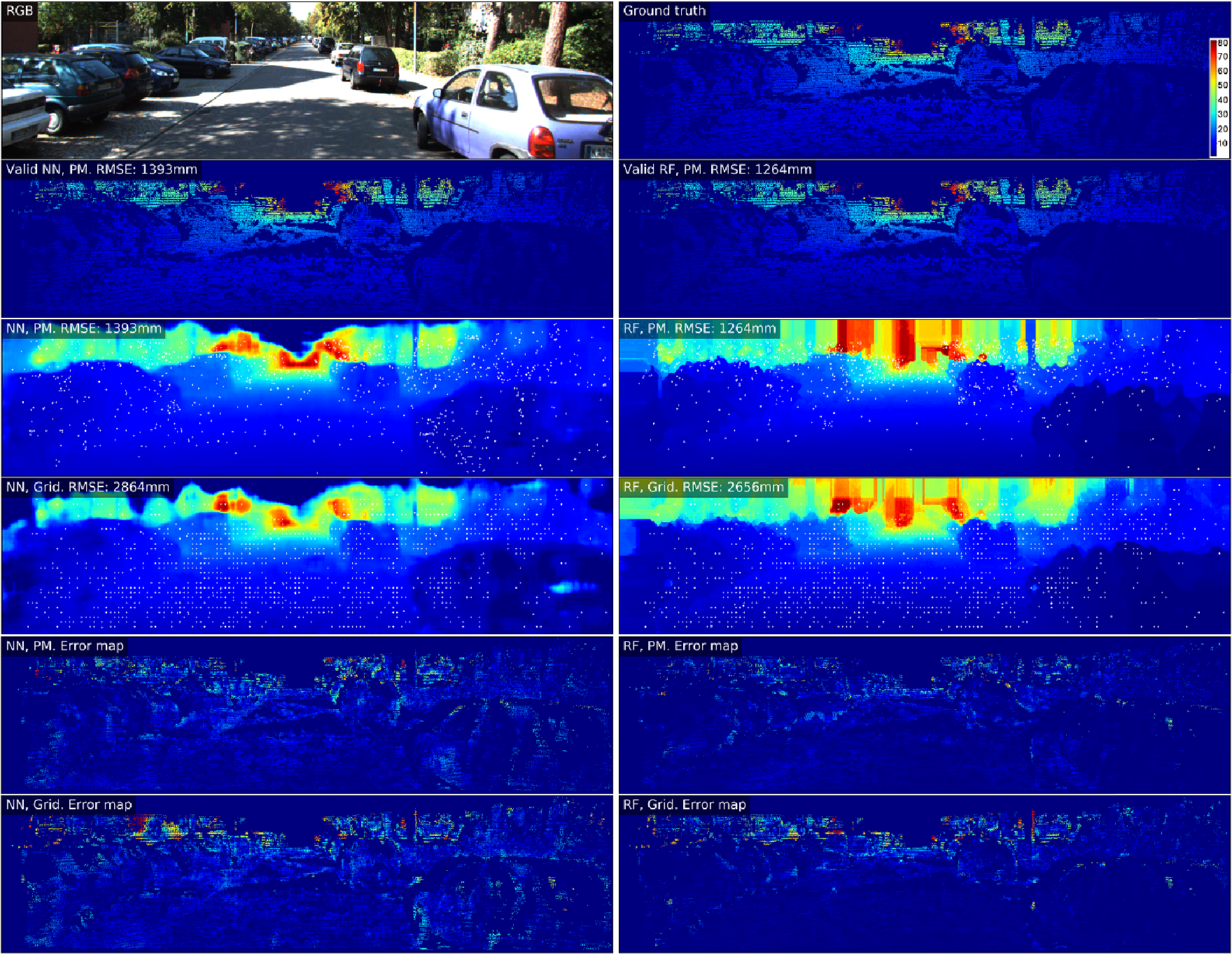}
\caption{Depth completion with the NN and RF completion methods combined with the PM and grid sampling methods (RGBd, 1024 samples). The first row shows the image and the ground truth (colors code for meters). The next three rows show the restriction of predictions with PM sampling to pixels with valid ground truth values, and the full predictions together with the sampling patterns (in white). The edges of objects are smoother when NN is employed compared to when RF is employed, because of the simplicity of the RF completion method, yet both perform well with PM in terms of RMSE. The last two rows show the error maps between the predictions and the ground truth. Large errors are shown in red, while small errors are in dark blue.} 
\label{fig:NN_RF_pred_compare}
\end{figure*}

\begin{figure}[tb]
\centering
\subfloat{%
       \includegraphics[width=3.49in]{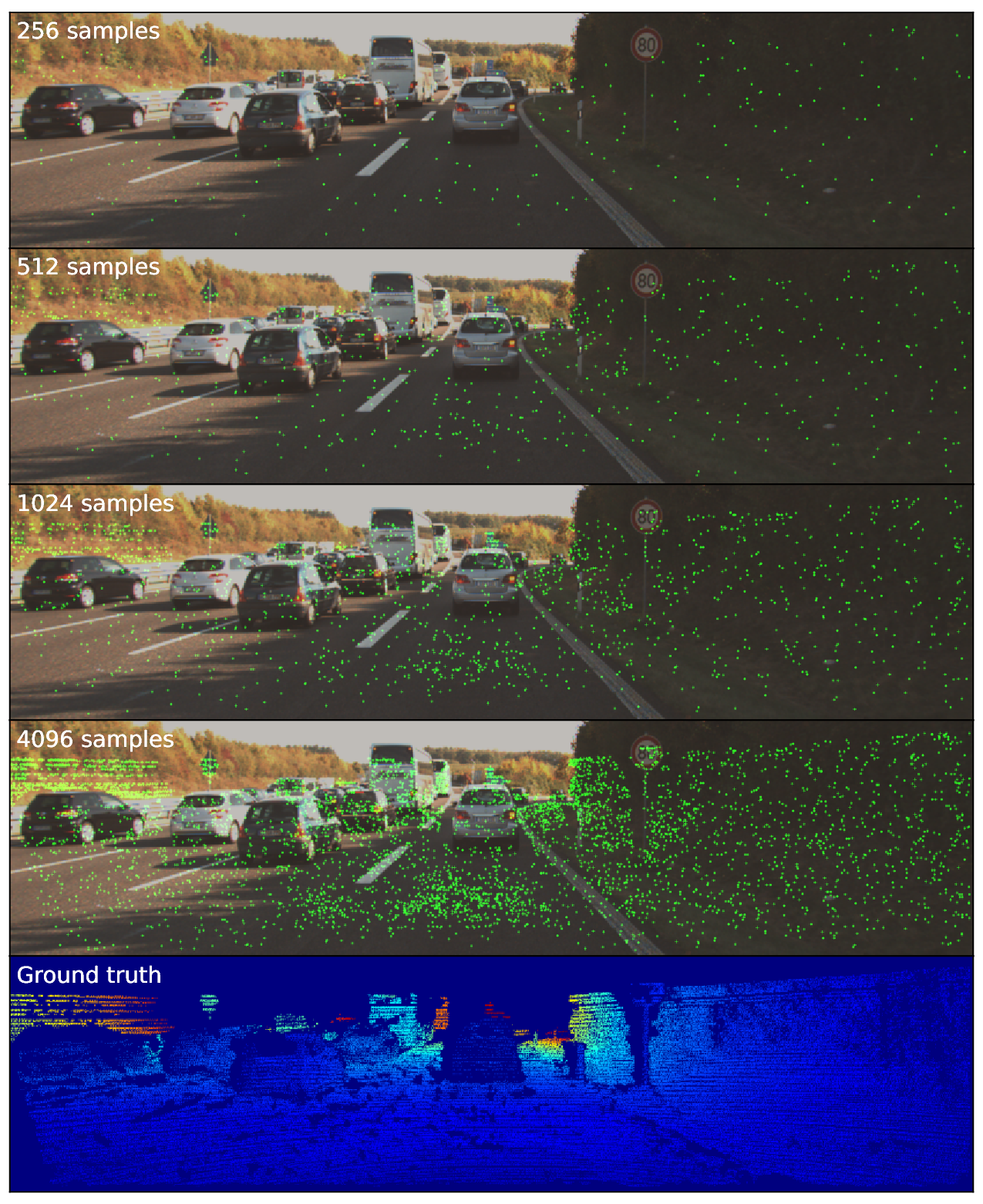}}
\caption{Final PM sampling patterns for various budgets with our NN-based construction. The non-uniform patterns reflect the variance-based sampling probabilities and also the available ground truth. As the sampling budget increases, denser sampling is performed in each phase.
} 
\label{fig:several_budgets}
\end{figure}

\subsection{A Neural Net-Based Implementation}\label{subsec:A Neural Net-Based Implementation}
The above generic construction accommodates any depth completion algorithm for which we can create different variants to form ensembles. As an almost straightforward implementation of this scheme we took the neural network (NN) algorithm of Ma et al. \cite{ma2018sparse,ma2019self} as our depth completion algorithm. It has an auto-encoder CNN architecture that maps optional RGB and mandatory sparse depth to dense depth, trained in a supervised manner. We created the variants by training it on distinct, roughly equal-sized subsets of the training set. To reduce overfitting, when calculating the variance for each pixel in the training stage, we excluded the predictor that was trained on that image. We note that this algorithm uses a validation set as part of its training, and this set was treated exactly like the test set. The same algorithm was also used for the final predictor. Beside the convenience of working with one, rather than two different algorithms, this choice intuitively increases the likelihood that the selected pixels would be suitable and useful for the final predictor.

\subsection{A Random Forest-Based Implementation}\label{subsec:A Random Forest-Based Implementation}
The Random Forest (RF) machine learning algorithm operates by training an ensemble of decision trees and predicting the value of a new sample by averaging their individual predictions \cite{Breiman2001,scikit-learn}. Each tree in this ensemble is constructed based on its own bootstrap sample of the training set, that is, a sample with replacement and of the same size as the original set. Our second implementation of the generic scheme involves a Random Forest predictor as the entire ensemble in each phase and also as the final predictor. 

More concretely, we formally use a decision-tree algorithm as the depth completion algorithm $A$ in the generic scheme. This algorithm treats depth completion as a regression problem for predicting the depth of a single pixel, and operates on feature vectors that will be described shortly. To train the ensemble in each phase, we first take a random subsample of the pixels of each training image, to reduce the size of the training sets of $A$. Then we create an ensemble by training $A$ using different bootstrap samples of that smaller training set. This is the same as training a Random Forest algorithm and using its internal ensemble. This very same Random Forest algorithm is also used as the final depth completion algorithm $A_f$. We comment that the size of the random subsample is 2048 pixels per image, which is a tiny fraction of the pixels in the images we used. We applied the ensemble predictors to all the pixels during training, including those they were trained on (see the generic training procedure in Figure~\ref{fig:generic algorithm}). 

The feature vector for a pixel $x$ is defined as follows. First, we convert RGB to HSV and take the color values and pixel coordinates as features. Except for the first phase, we then find the three nearest measured pixels (in the $L_1$ distance) and engineer several additional features. These include, for each neighbor, its measured depth, its $L_1$ distance from $x$, and for each coordinate and color value, the difference from the respective value for $x$. We thus represent each pixel as a $5+ 7 \times 3 = 26$-dimensional feature vector. For the depth-only d scenario, the same construction holds without the color-related features, yielding 14 features. 

Figures \ref{fig:NN_RF_pred_compare} and \ref{fig:several_budgets} demonstrate the RF-based and NN-based implementations as applied to real data.

\subsection{Probability Matching}
Focusing measurement on high-variance pixels is motivated by the view of variance as a proxy for error. Specifically, regarding the prediction of an ensemble member on a pixel as a random variable $P$, and denoting $g$ for the ground truth value, we have by the bias-variance trade-off that
\begin{equation}
\E[(P-g)^2] = Var(P) + (\E[P]-g)^2\;.
\end{equation}
For an idealized ensemble of infinite size and i.i.d. members, the squared error of the ensemble is lower bounded by the variance of the predictions. Furthermore, if the predictor has a relatively small bias, the variance approximates the squared error. We note that in our implementation the ensemble is finite, so expectations are approximated by finite sums, and that in our NN-based implementation the predictor variants are not i.i.d. but based on distinct subsets.

Regarding variance as a proxy for error, we may ask what is the optimal way for reducing the total variance in an image given a budget of $B$ pixels to measure. A solution to this question seems infeasible for a complex predictor, let alone a black-box one, but we may analyze a simplified scenario. 

Consider an image with $n$ pixels and assume that a single pixel needs to be selected. We will make the further simplifying assumption that selecting a pixel reduces its variance to zero, without affecting the variance of all the other pixels. The utility of selecting pixel $i$, denoted $u_i$, is thus simply the variance. The optimal strategy for maximizing the utility, denoted MAX, trivially selects the best pixel. Its utility thus satisfies $u_{MAX}=\max_i\{u_i\}$.

Our probability matching strategy, denoted PM, selects a pixel with probability proportional to its variance, and has expected utility 
\begin{equation}
u_{PM} = \sum_{i=1}^n\frac{u_i}{\sum_{j=1}^n u_j} \cdot u_i = \frac{\sum_{i=1}^n u_i^2}{\sum_{i=1}^n u_i}\;.
\end{equation}
Finally, a strategy of completely random choice, denoted RND, has expected utility $u_{RND} = (1/n)\sum_{i=1}^n u_i$.

The expected utility of PM is always better than that of RND except for trivial scenarios. We have that 
\begin{align}
u_{PM} &= \frac{\sum_{i=1}^n u_i^2}{\sum_{i=1}^n u_i} \geq \left(\frac{1}{n}\sum_{i=1}^n u_i^2\right)^\frac{1}{2} \geq \frac{1}{n}\sum_{i=1}^n u_i \nonumber\\ &= u_{RND}\;,
\end{align}
where both inequalities follow from the inequality of the quadratic and arithmetic means, and equality is possible only if $u_1=\ldots=u_n$.
Furthermore, 
\begin{equation}
u_{PM} \geq \left(\frac{1}{n}\sum_{i=1}^n u_i^2\right)^\frac{1}{2}\geq n^{-\frac{1}{2}}\max_i\{u_i\} = n^{-\frac{1}{2}} u_{MAX}\;,
\end{equation}
compared with $u_{RND}$ which may be as low as $n^{-1}u_{MAX}$, specifically if only one pixel has non-zero variance. Finally, we have that
\begin{equation}
u_{PM} \geq \frac{\max_i\{u_i^2\}}{\sum_{i=1}^n u_i} = \frac{\max_i\{u_i\}}{\sum_{i=1}^n u_i} \cdot u_{MAX} \;,
\end{equation}
so if the utility of one pixel dominates the sum, $u_{PM}$ would get arbitrarily close to $u_{MAX}$, while $u_{RND}$ would not. In particular, if there is a single positive $u_i$, \begin{equation}
u_{PM} = u_{MAX} = n\cdot u_{RND}\;.
\end{equation}
The above properties justify theoretically why PM is superior to RND, a phenomenon that is also observed in our experiments (see Section \ref{sec:Experiments and Results}). However, these experiments show that PM is also superior to the theoretically optimal MAX, which empirically does even worse than RND. It appears that for a normal-sized budget, the greedy approach of MAX causes it to spend its budget redundantly in small areas of the image, while PM inherently explores different parts of it. From a so-called \textit{exploration-exploitation} perspective, it appears that MAX focuses too much on exploitation, RND by definition focuses only on exploration, and PM strikes the best balance. These observations, however, are outside our theoretical analysis.
\section{Experiments and Results}\label{sec:Experiments and Results}

We studied two different implementations of our adaptive sampling method using the KITTI depth completion dataset \cite{Uhrig2017THREEDV}. 
In one implementation, the supervised NN-based depth completion algorithm of Ma et al. was used as a black box in our construction. We also applied our method in conjunction with our Random Forest-based depth completion algorithm.

\subsection{Data Preparation}\label{subsec:Data Preparation}

The KITTI dataset includes a training set of 85898 frames (divided into 138 different sequences, or \textit{drives}) and 1000 selected validation frames, as well as other data. These scenes have a maximal depth of approximately 85 meters. Each frame has an RGB image, a corresponding noisy, sparse LiDAR scan, and an enhanced, semi-dense depth map (the \textit{ground truth}). While the raw scans have an average of about 5\% of the pixels annotated for depth, the ground truth, which is enhanced based on several raw scans and stereo images, has an average of about 15\% annotated pixels. We use the ground truth data as the source of depth information in our experiments. This choice serves to better test the ability of algorithms to freely select sampling points based on their potential benefit.

We used the 31 smallest drives in the full KITTI training set for the purpose of training our NNs. This subset, which comprises 11994 images, is diverse and comprehensive enough for the task, and its smaller size helped reduce run-time. 

To train our RF-based implementation, this set was further reduced to 500 images to accommodate computational constraints. The images of this subset were selected by skipping a fixed number of images in each drive while alternating between the right and left cameras. 

For validation and testing, we split the KITTI selected validation set into 203 validation images and 797 test images. Diverse scenes (both urban and rural) were allocated to each subset, with images from the same drive always being allocated to the same subset. To match the 1216$\times$352 dimensions of validation and test images, a similarly-sized area was cropped from the bottom and center of every training image.

\subsection{Algorithmic Settings}\label{subsec:Algorithmic Settings}
Our NN-based implementation uses the algorithm and code of Ma et al. \cite{ma2018sparse,ma2019self}. We used their supervised algorithm with a batch size of 4, an 18-layer architecture, and a maximum of 7 training epochs instead of the default 11, to save run-time. Otherwise, their default settings were used. For our method, we used 4 phases and ensembles of size 5. For training ensemble NNs, the drives in the training set were divided into five subsets, which were roughly balanced in terms of image counts. This implementation can be found on GitHub.\footnote{\url{https://github.com/shacharp/Adaptive-LiDAR-Sampling}}

The RF-based method was implemented using scikit-learn \cite{scikit-learn}. The number of trees per forest was 40, except for the final predictor where the number of trees was 500. Otherwise, the default settings for random forest regression were used. Our method was run with 8 phases. 

As an example of runtimes, with grid or random sampling, the NN-based method required $\sim$100 milliseconds per image for inference using the underlying neural net. In comparison, for PM sampling, $\sim$1100 milliseconds per image were required. These times were measured while running on two NVIDIA GeForce RTX 2080 Ti GPUs. We note that parallelizing the ensemble computations (see Subsection \ref{subsec:generic algorithm}) and using simpler and faster ensemble members for calculating variance could bring the running time of PM much closer to that of random or grid.

\subsection{Evaluation Metrics}\label{subsec:Evaluation Metrics}
In our empirical analysis we report the root mean squared error (RMSE) and the mean absolute error (MAE), computed over all test pixels with annotated depth, as well as the absolute relative error (REL) and the $\delta_1$ measure. These last two metrics are defined by
\begin{equation}
\mathrm{REL} = (1/N)\sum_{i=1}^N |p_i-g_i|/g_i
\end{equation}
and 
\begin{equation}
\delta_1 = (1/N)\sum_{i=1}^N\Ind{\max\{p_i/g_i,g_i/p_i\}< 1.25}\;,
\end{equation}
where $\{(p_i,g_i)\}_{i=1}^N$ are pairs of predicted depths and ground truth
values, and $\Ind{E}$ denotes the indicator function of an event $E$. We note that both the NN-based and RF-based depth completion algorithms as well as the PM sampling method are geared towards optimizing the RMSE. Thus, we expect mostly this criterion to be affected by our techniques. 

\subsection{Our Results}\label{subsec:Our Results}
We tested our methods using the KITTI data described in Subsection \ref{subsec:Data Preparation}. As the core depth completion algorithm we used either the algorithm of Ma et al. (NN) or our own Random Forest-based algorithm (RF). As the sampling method we tried random and grid sampling, our phased ensemble-based sampling with probability matching (PM) and with the greedy maximal variance choice (MAX), as well as the super-pixel sampler of Wolff et al. \cite{wolff2020icra}. We emphasize that for each different experimental setting (combining completion method, sampling method, budget, data type, etc.), we performed a separate training process, yielding a predictor, and where relevant, also ensembles, optimized for that setting. It should be noted that grid sampling had to be approximated owing to the semi-dense nature of the ground truth. Also included in the comparison is the end-to-end adaptive sampling and reconstruction method of Bergman et al. \cite{bergman2020deep}. 

\begin{figure}[htbp]
\centering
\includegraphics[width=3.49in]{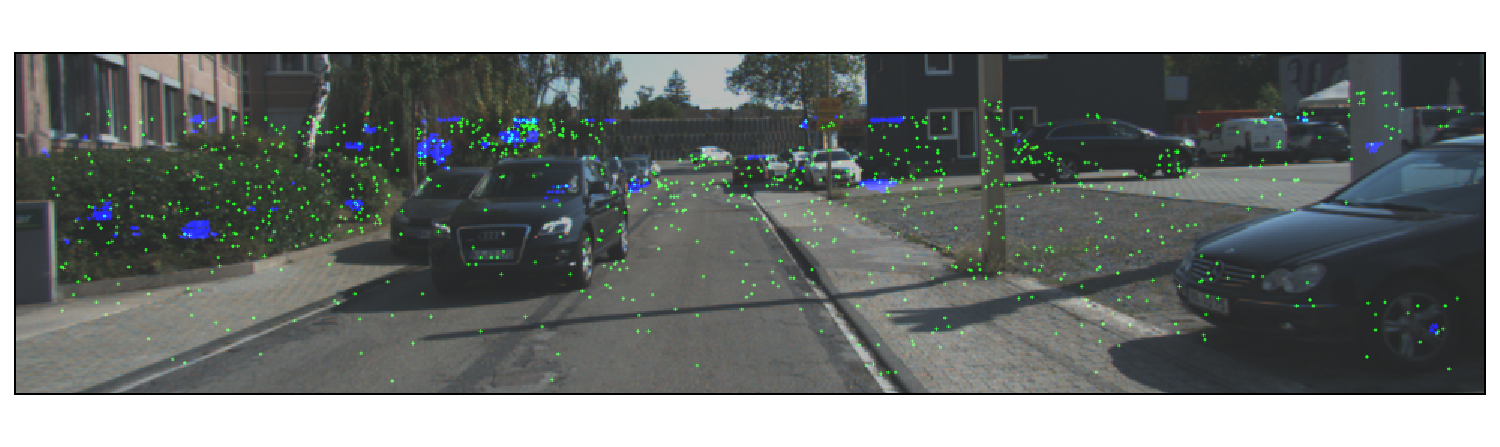}
\caption{Sampling patterns for PM (green) and MAX (blue) with our NN-based construction. While PM inherently balances exploration and exploitation, MAX does not, leading to redundant sampling in limited regions.}
\label{fig:pm_vs_max}
\end{figure}

The results for the RGBd setting with a budget of 1024 samples are summarized in Table~\ref{tab:kitti_metrics_1024}.
\begin{table}[ht]
\caption{Performance on the KITTI subset (RGBd, 1024 samples)}
\label{tab:kitti_metrics_1024}
\centering
    \begin{threeparttable}
    \begin{tabular}{|c|c|c|c|c|c|}
    \hline
    Completion & Sampling & RMSE & MAE & \multirow{2}{*}{REL} & \multirow{2}{*}{$\delta_1$ [$\%$]} \\ 
      Method & Method & [mm] & [mm] & & \\
    \hline
    \multirow{5}{*}{NN} & Random & 1737 & 655 & 0.038 & 98.45\\
     & Grid & 1825 & 688 & 0.040 & 98.29\\
     & Ours (PM) & \textbf{1077} & \textbf{473} & \textbf{0.030} & 99.24\\
     & MAX & 3777 & 2090 & 0.138 & 82.25\\
     & Wolff et al. \cite{wolff2020icra} & 1302 & 522 & \textbf{0.030} & \textbf{99.30}\\
    \hline
    \multirow{5}{*}{RF} & Random & 2357 & 666 & 0.032 & 97.99 \\
     & Grid & 2348 & 647 & 0.031 & 98.06\\
     & Ours (PM) & \textbf{1092} & \textbf{414} & 0.027 & \textbf{99.27}\\
     & MAX & 2308 & 1112 & 0.081 & 92.81\\
     & Wolff et al. \cite{wolff2020icra} & 1645 & 489 & \textbf{0.022} & 99.12\\
    \hline
    \multicolumn{2}{|c|}{Bergman et al. \cite{bergman2020deep}\tnote{1}} & $\sim$1600 & -- & -- & --\\
    \hline
    \end{tabular}
    \begin{tablenotes}
    \item[1] The comparison with this method is approximate.
  \end{tablenotes}
    \end{threeparttable}
\end{table}
It can be seen that the PM sampling method is the best one in conjunction with both completion methods, despite competition from the method of Wolff et al. for the REL and $\delta_1$ relative measures. The greedy MAX method does even worse than random and grid, a fact explained by its highly redundant sampling pattern (Figure~\ref{fig:pm_vs_max}). 
\begin{figure}[htbp]
\centering
\includegraphics[scale=0.12]{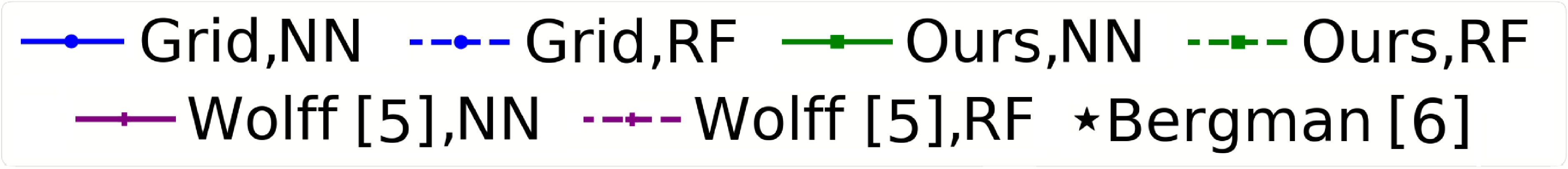}
\subfloat{%
       \includegraphics[scale=0.085]{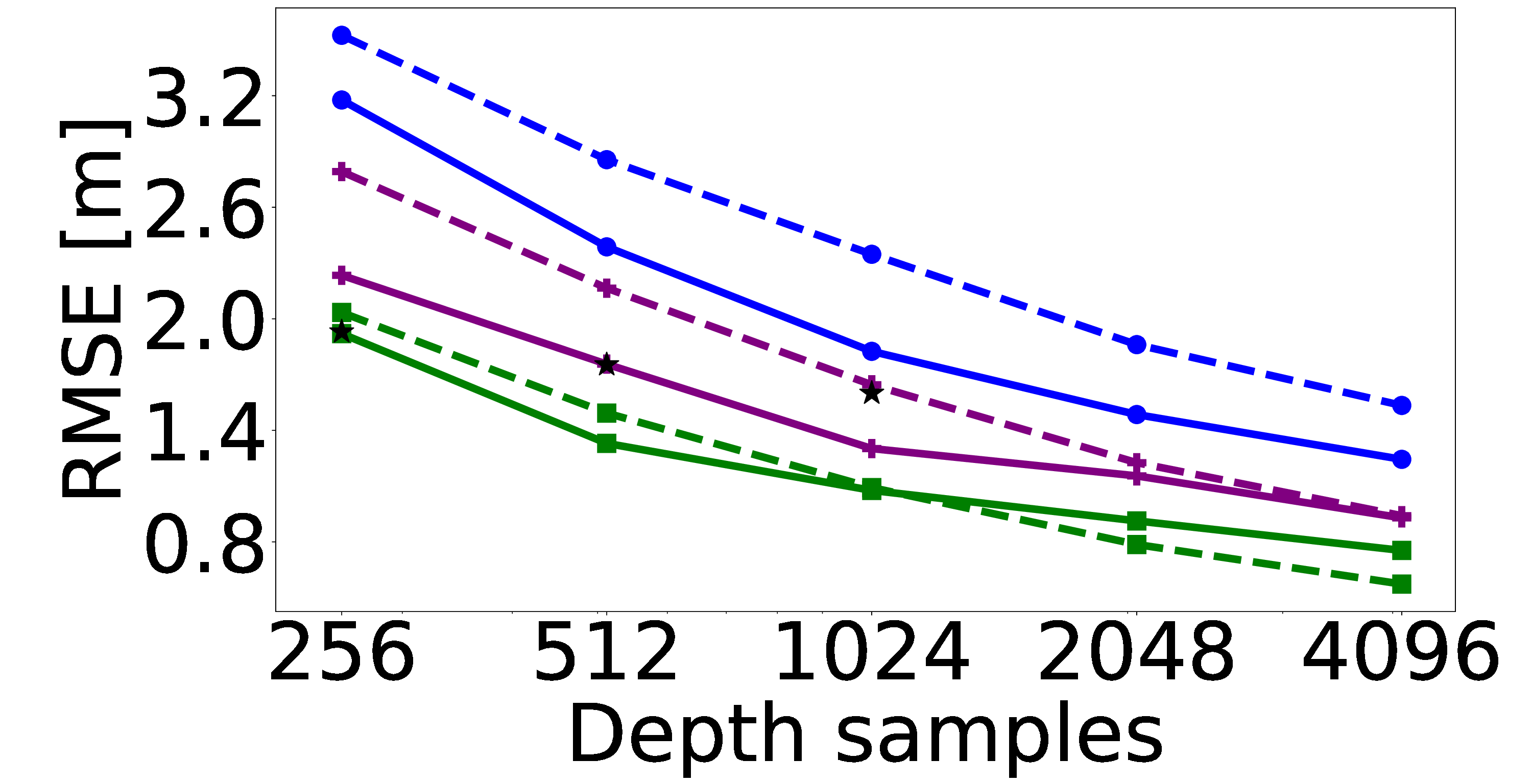}}
\hfil
\subfloat{%
       \includegraphics[scale=0.085]{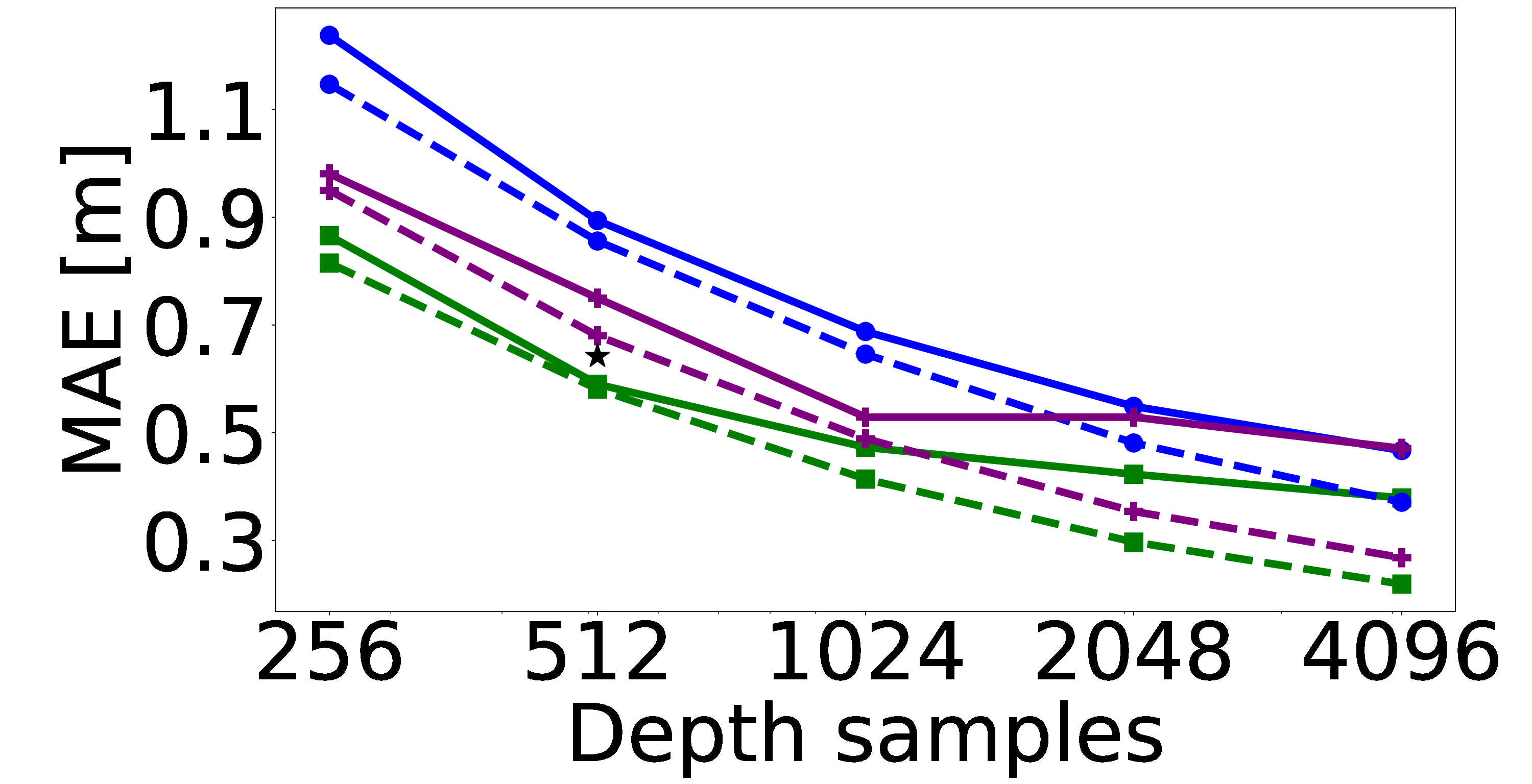}}
\hfil
\subfloat{%
       \includegraphics[scale=0.085]{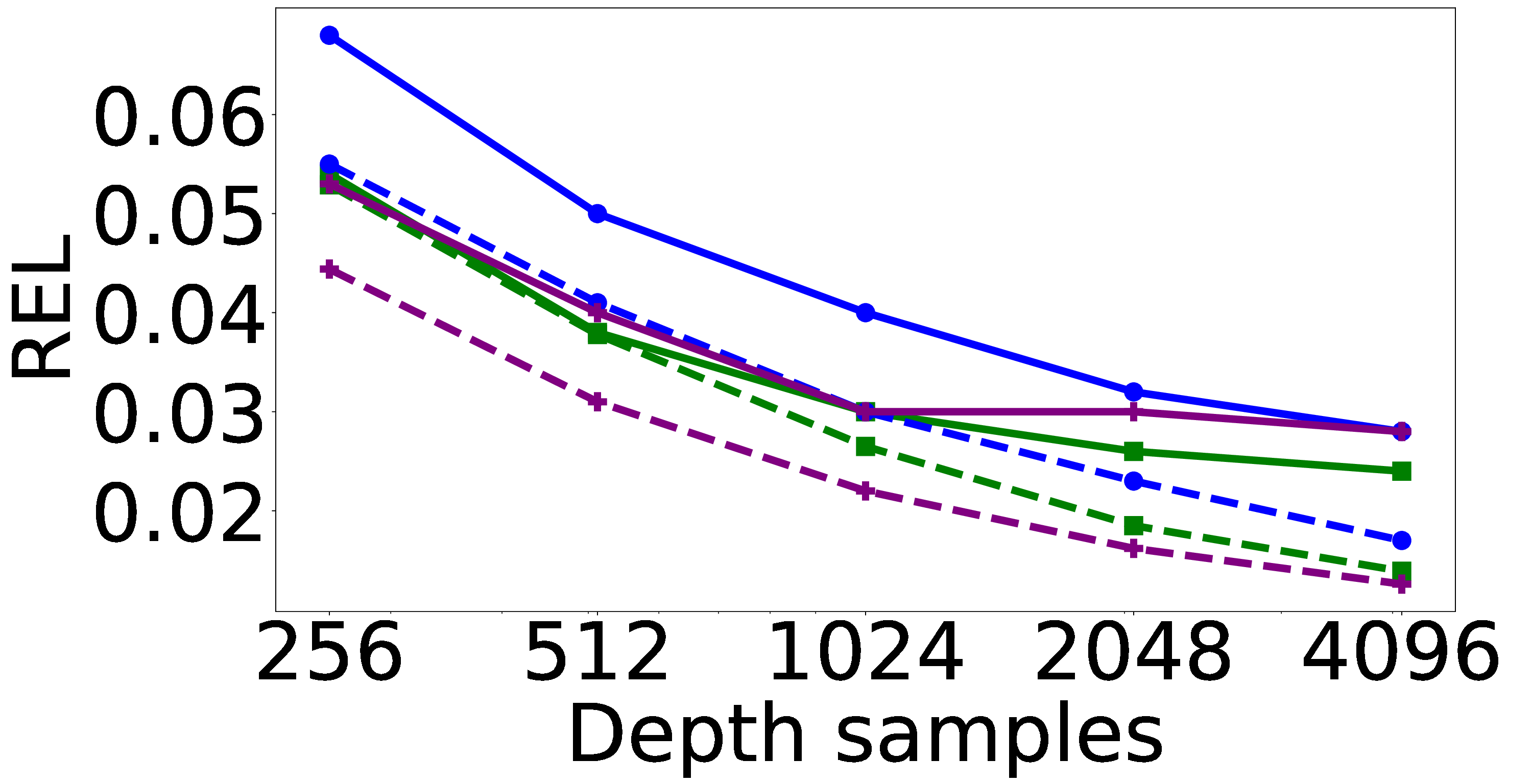}}
\hfil
\subfloat{%
       \includegraphics[scale=0.085]{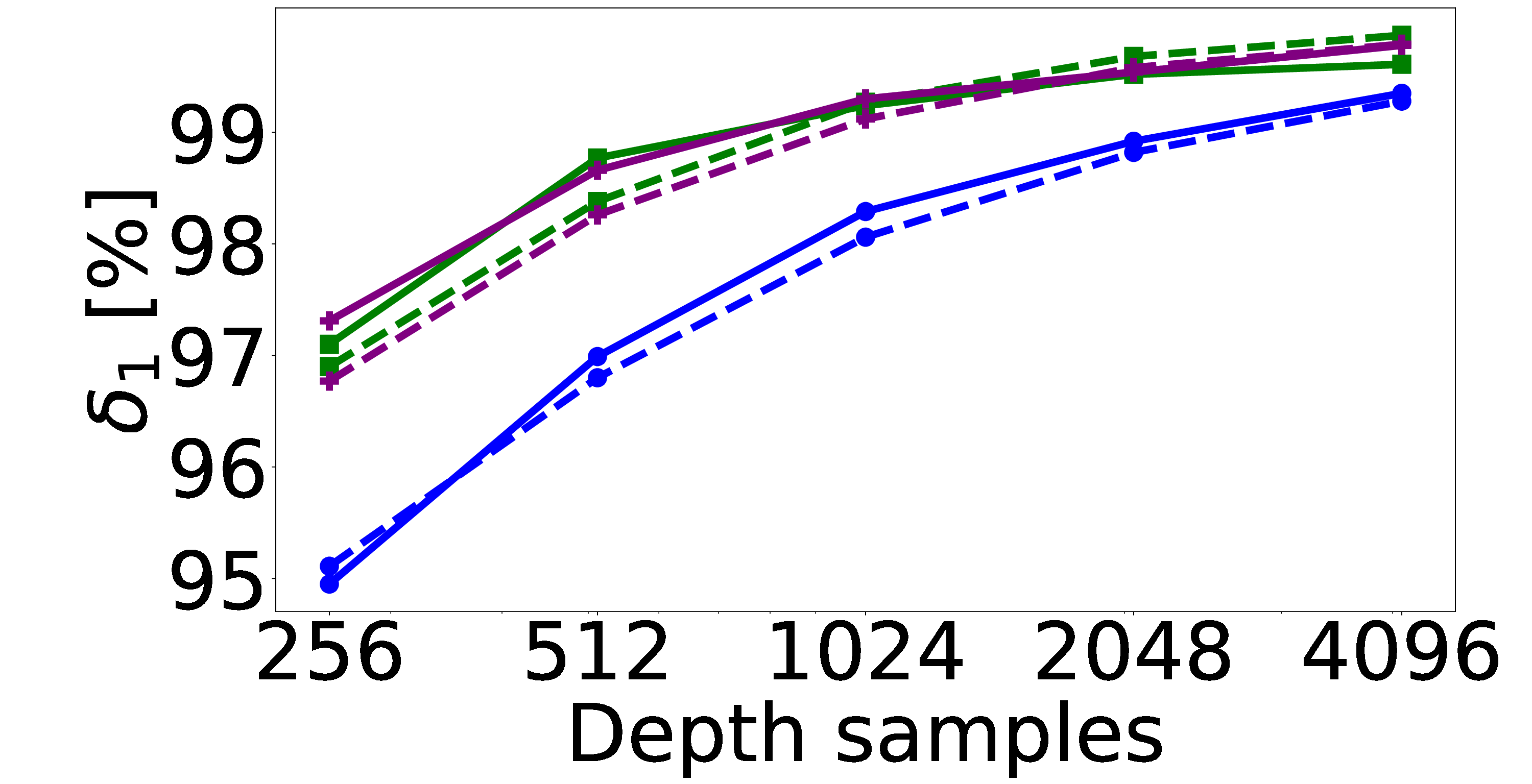}}
\caption{Prediction accuracy with different sampling budgets on the KITTI subset with RGB and depth data. The respective sparsity levels are 0.06\%, 0.12\%, 0.24\%, 0.48\%, and 0.96\% of all image pixels. Results reported by Bergman et al. \cite{bergman2020deep} are given as a point of reference.}
\label{fig:kitti_metrics_sparsity_rgbd_combined_no_rand}
\end{figure}
\begin{figure}[htbp]
\centering
\includegraphics[scale=0.12]{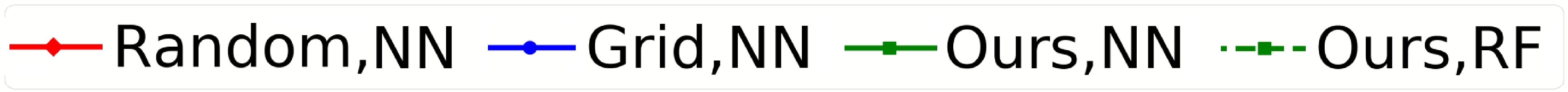}
\subfloat{%
       \includegraphics[scale=0.085]{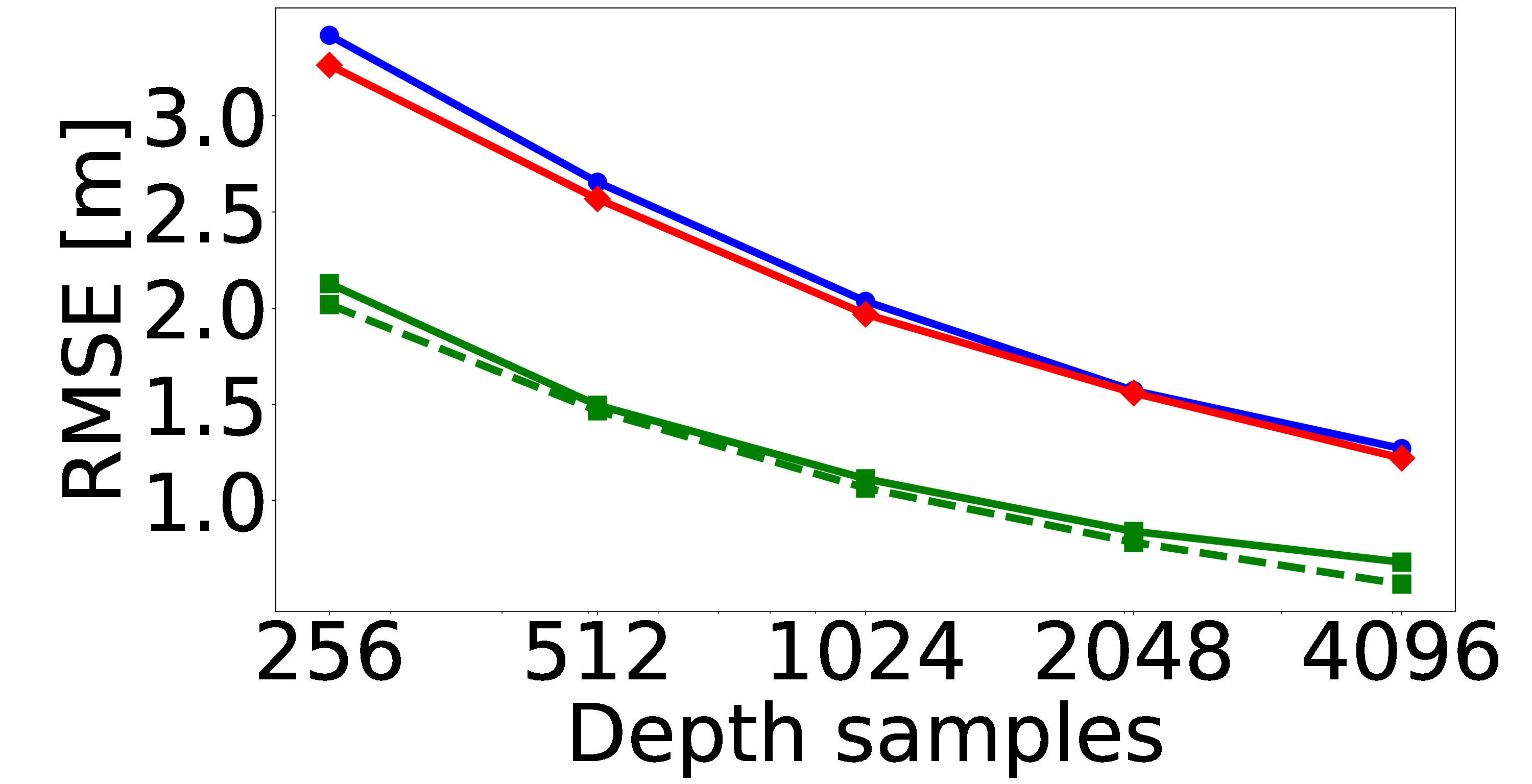}}
\hfil
\subfloat{%
       \includegraphics[scale=0.085]{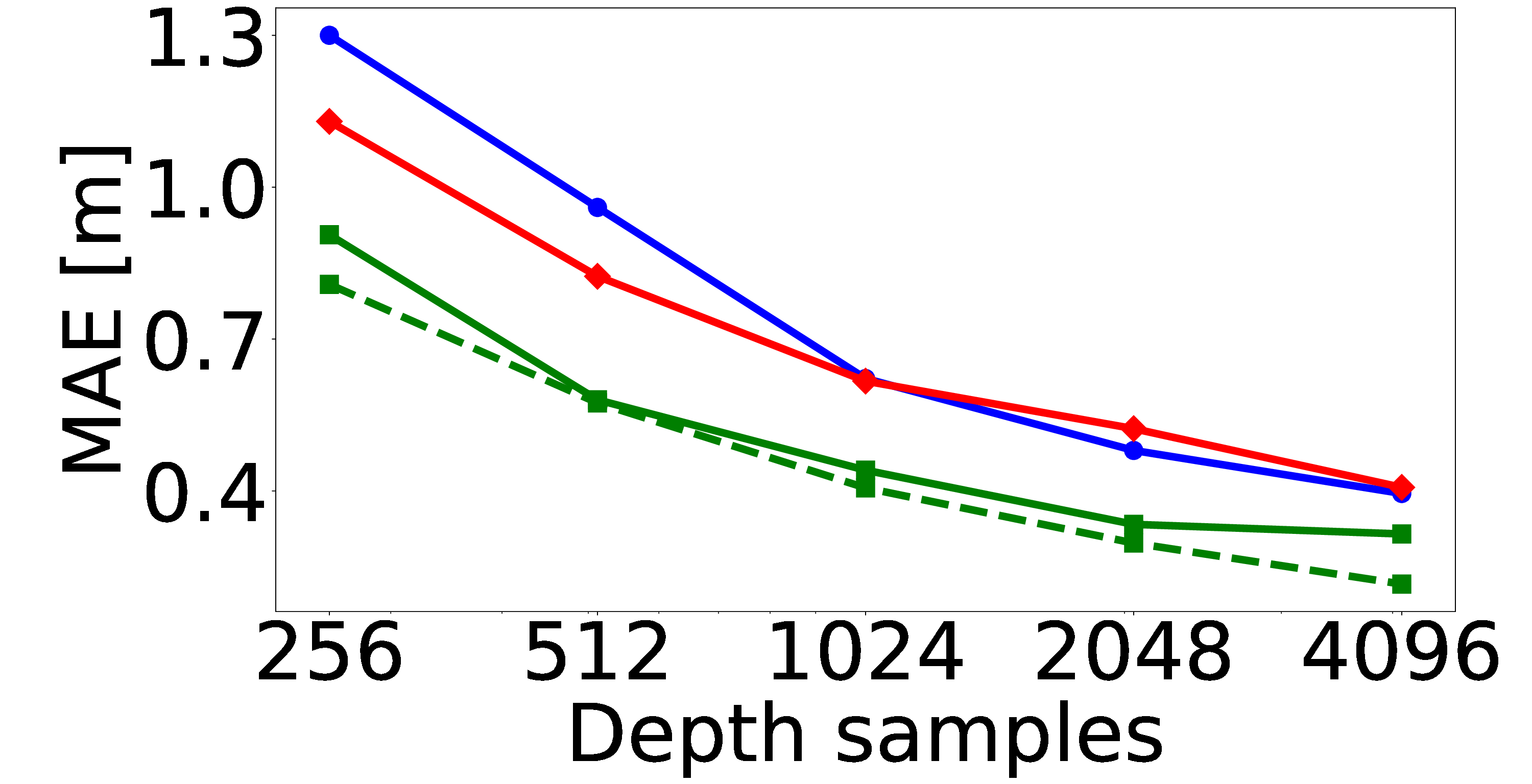}}
\hfil
\subfloat{%
       \includegraphics[scale=0.085]{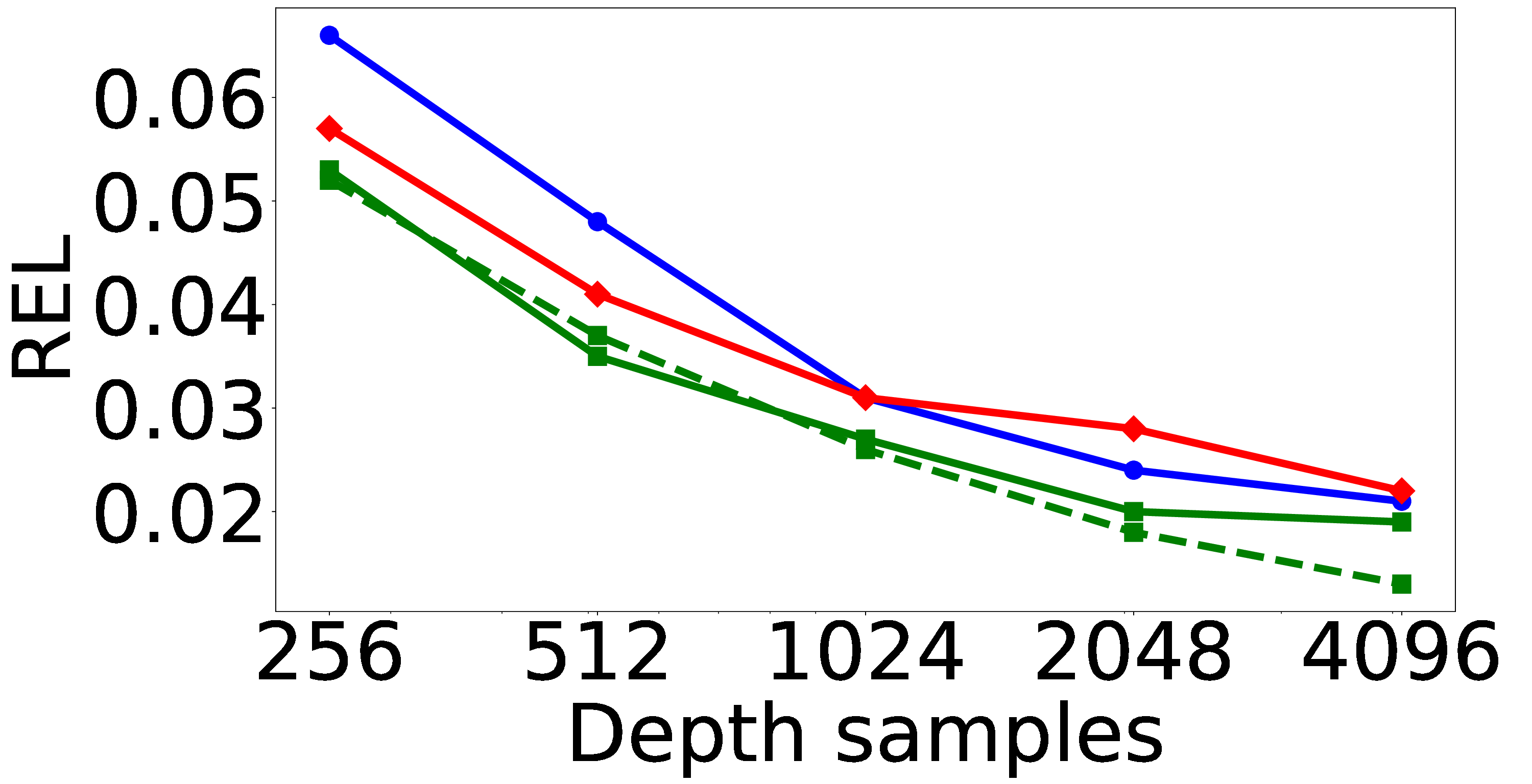}}
\hfil
\subfloat{%
       \includegraphics[scale=0.085]{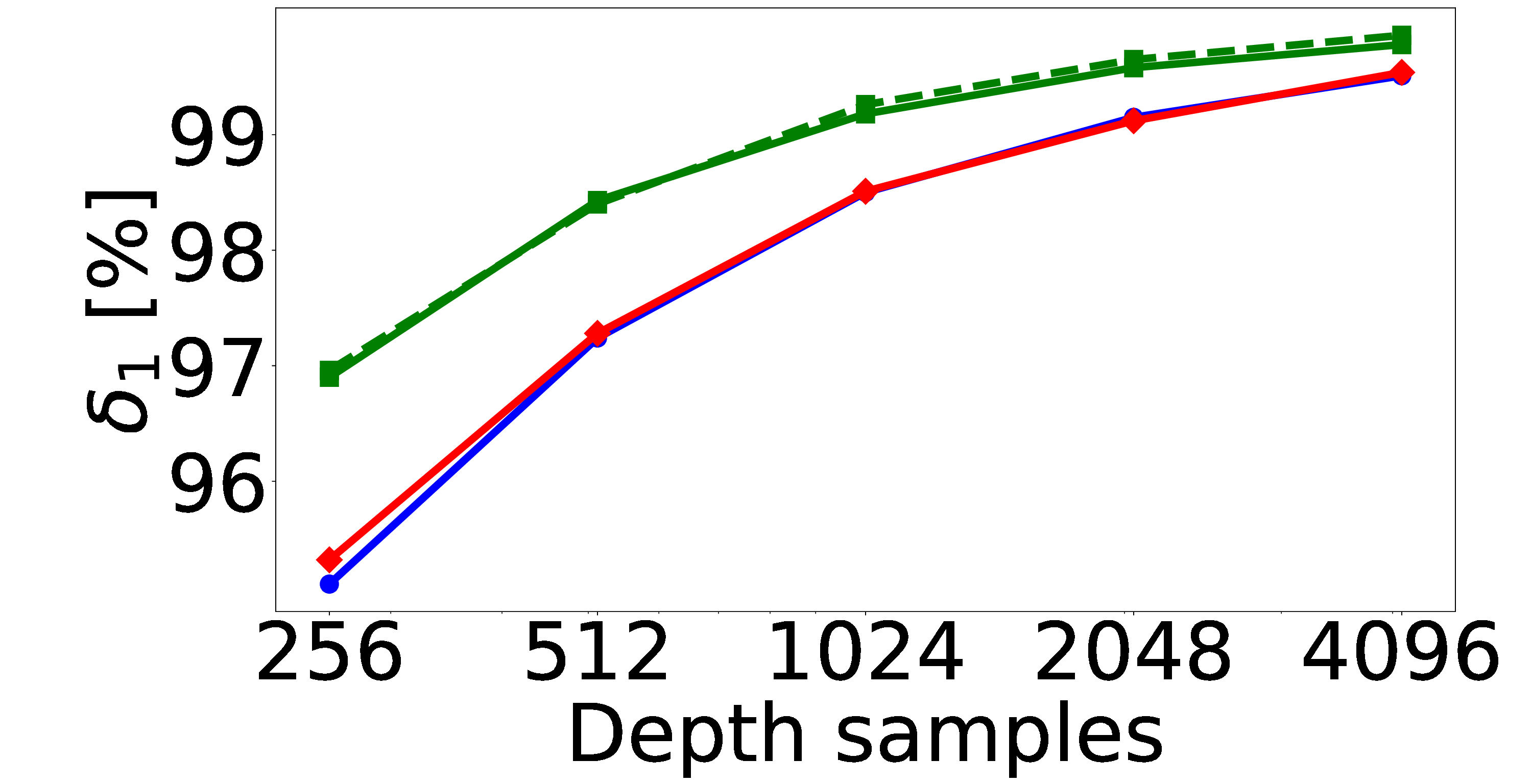}}
\caption{Prediction accuracy with depth data alone given different sampling budgets on the KITTI subset.%
}
\label{fig:kitti_pm_error}
\end{figure}

\begin{figure}[htbp]
\centering
\includegraphics[width=3.49in]{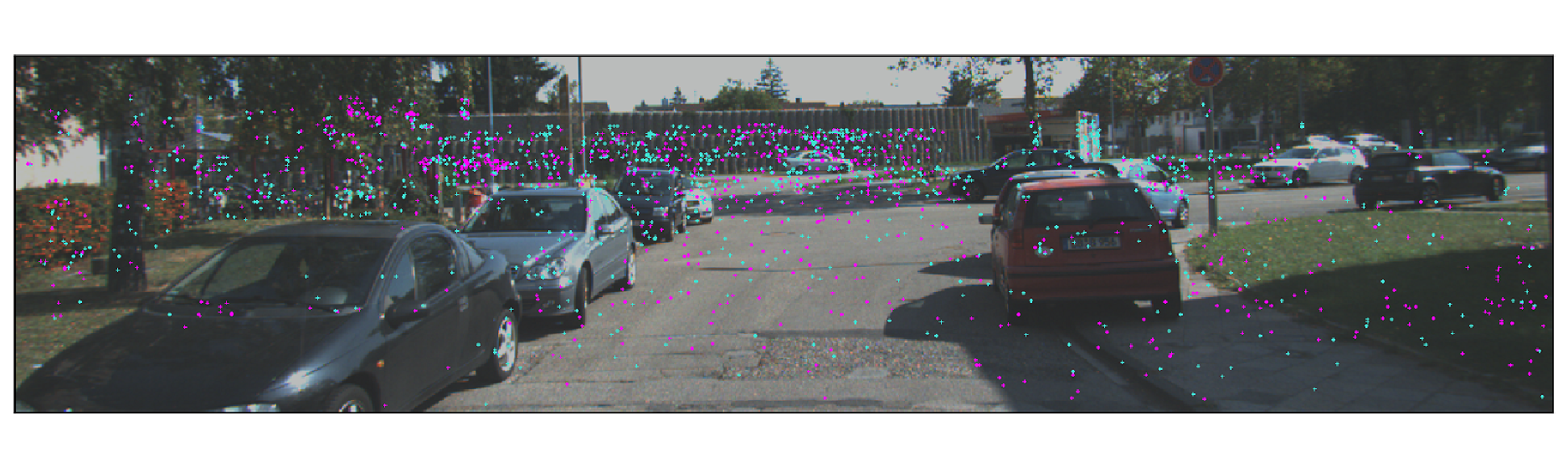}
\caption{Sampling patterns for our NN-based method with RGBd data (turquoise) and depth-only data (magenta), with a sampling budget of 1024.}
\label{fig:RGBd_vs_d_sampling_patterns}
\end{figure}

\begin{figure}[htbp]
\centering
\includegraphics[width=3.49in]{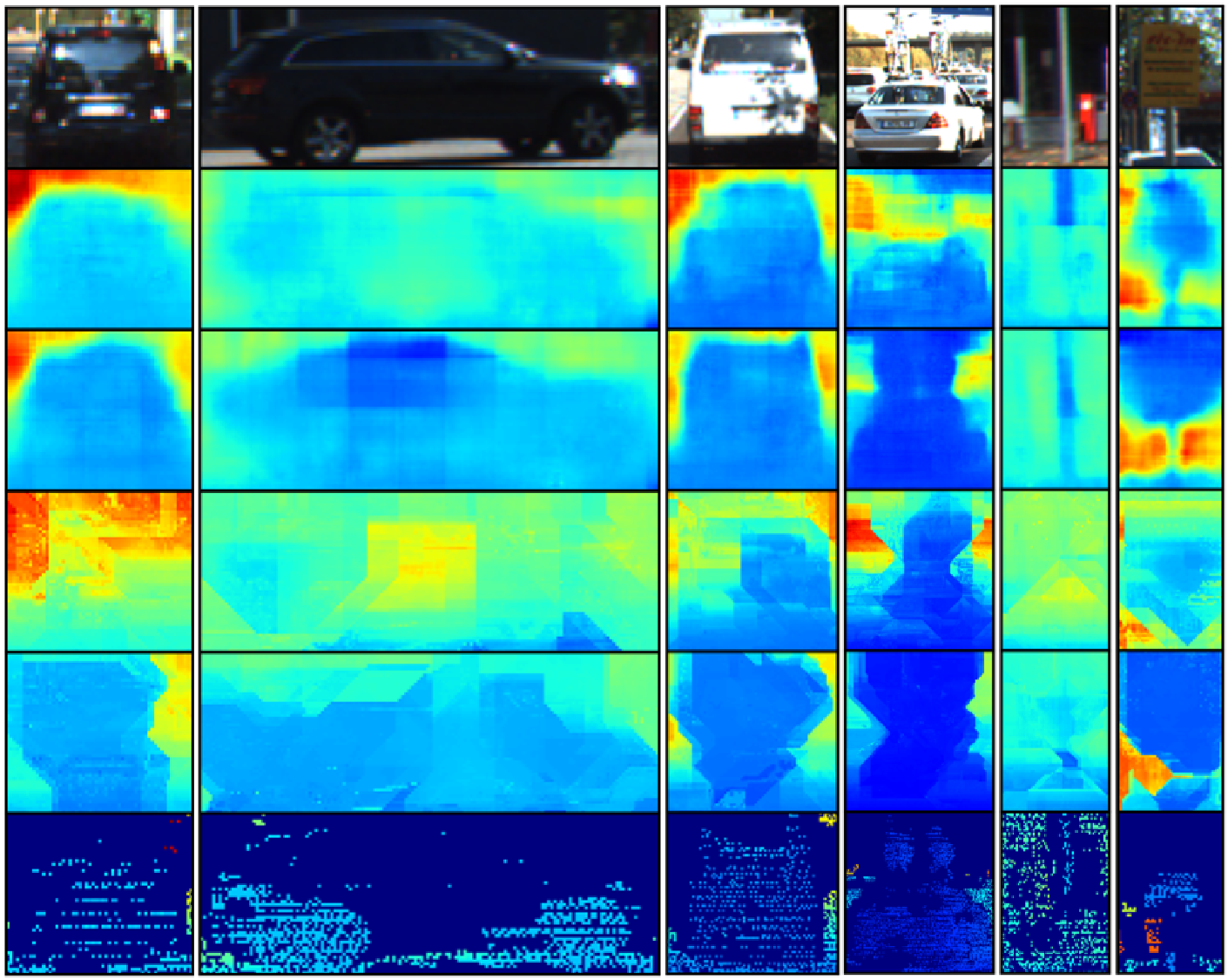}
\caption{Qualitative depth completion results on KITTI, with RGBd data and a sampling budget of 256, or 0.06\% of all image pixels. Top to bottom: RGB, NN grid sampling, NN PM sampling, RF grid sampling, RF PM sampling, and ground truth. Object shapes and interiors are better preserved with PM. For RF, the edges are generally rougher than NN, yet the advantage of PM over grid is even more pronounced.}
\label{fig:close_up}
\end{figure}
We note that the comparison with the method of Bergman et al. is only approximate, since they use the entire KITTI training data and report results on the entire validation set. With this caveat, our PM method gives superior results for both completion methods.

The advantage of our PM method over non-adaptive grid sampling across sampling budgets is shown in Figure~\ref{fig:kitti_metrics_sparsity_rgbd_combined_no_rand}. 
Both the NN and the RF completion methods enjoy better performance in conjunction with PM, compared to grid in all measures. As the budget increases, the advantage understandably decreases, but a bit less so for RF. Compared to other adaptive sampling methods, our algorithms are better for RMSE (which the net and RF optimize for) and MAE. We found that the classical sampling of 
\cite{wolff2020icra} yields good results for REL. Visual comparisons of PM and grid sampling are shown in Figures \ref{fig:NN_RF_pred_compare} and \ref{fig:close_up}. 
Another way to measure the quality of an adaptive depth completion algorithm is by the fraction of pixels one needs to sample to obtain a given RMSE goal, as shown in Table~\ref{tab:kitti_fixed_nn}.
\begin{table*}[htbp]
\caption{Sampling budget (as \% of total pixels) required to obtain target performance} 
\label{tab:kitti_fixed_nn}
\centering
  \begin{threeparttable}
    \begin{tabular}{|c|c|c|c|c|c|c|}
    \hline
    Completion & Sampling & 
    \multicolumn{5}{c|}{RMSE [mm]}\\ 
    \cline{3-7}
    Method & Method & 1000 & 1250 & 1500 & 1750 & 2000 \\ 
    \hline
    NN & Random & 1.609\% & 0.762\% & 0.413\% & 0.247\% & 0.158\% \\ 
    NN & Grid & 1.822\% & 0.926\% & 0.532\% & 0.333\% & 0.222\% \\ 
    NN & Ours (PM) &     0.389\% & 0.188\% & 0.103\% & 0.062\% & 0.040\% \\  
    NN & Wolff et al. \cite{wolff2020icra} & 0.550\% & 0.274\% & 0.155\% & 0.096\% & 0.063\% \\ 
    \hline
    \multicolumn{2}{|c|}{Bergman et al. \cite{bergman2020deep}} & [3.022\%]\tnote{$\dagger$} & [0.847\%] & [0.300\%] & 0.124\% & 0.058\% \\
    \hline
    \multicolumn{2}{|c|}{\textbf{PM vs. Grid Ratio (NN)}} & 1:4.7 & 1:4.9 & 1:5.2 & 1:5.4 & 1:5.5 \\ 
    \hline
    
    RF & Random & 2.635\% & 1.388\% & 0.822\% & 0.528\% & 0.360\% \\ 
    RF & Grid & 3.022\% & 1.545\% & 0.893\% & 0.561\% & 0.376\% \\ 
    RF & Ours (PM) & 0.284\% & 0.175\% & 0.118\% & 0.085\% & 0.063\% \\ 

    RF & Wolff et al. \cite{wolff2020icra} & 0.820\% & 0.467\% & 0.295\% & 0.200\% & 0.143\% \\ 
    \hline
    \multicolumn{2}{|c|}{\textbf{PM vs. Grid Ratio (RF)}} & 1:10.6 & 1:8.8 & 1:7.6 & 1:6.6 & 1:5.9 \\

    \hline
    \end{tabular}
    \begin{tablenotes}
    \item[$\dagger$]Bracketed expressions are extrapolated.
    \end{tablenotes}
  \end{threeparttable}
\end{table*}
In addition, we give the ratio between the budgets required with the PM method and with grid sampling for both the RF and NN completion methods. It can be seen that our methods allow for significant savings in sampling, with some dependence on the required accuracy. The advantage of using PM over grid sampling is particularly noticeable for the RF-based completion method. We note that the numbers in Table~\ref{tab:kitti_fixed_nn} are interpolated from available results by linear regression of the logarithm of the budget against the logarithm of the RMSE, and for the method of Bergman et al., they are also extrapolated.

As noted before, our methods are also applicable when only depth data is available. Results for this setting are given in Figure~\ref{fig:kitti_pm_error}, which shows the same general behavior as with RGBd data. We also note that having RGB data in addition to depth helps performance moderately for small budgets but this effect decreases and may even reverse for larger budgets. This phenomenon is explained by the fact that concrete depth measurements are much more informative than RGB data, and that at some point, RGB data carries little additional relevant information considering its huge size. An example comparing the sampling patterns with RGBd data and with depth alone is given in Figure~\ref{fig:RGBd_vs_d_sampling_patterns}.

\begin{figure}[htbp]
\centering
\subfloat{%
       \includegraphics[scale=0.35]{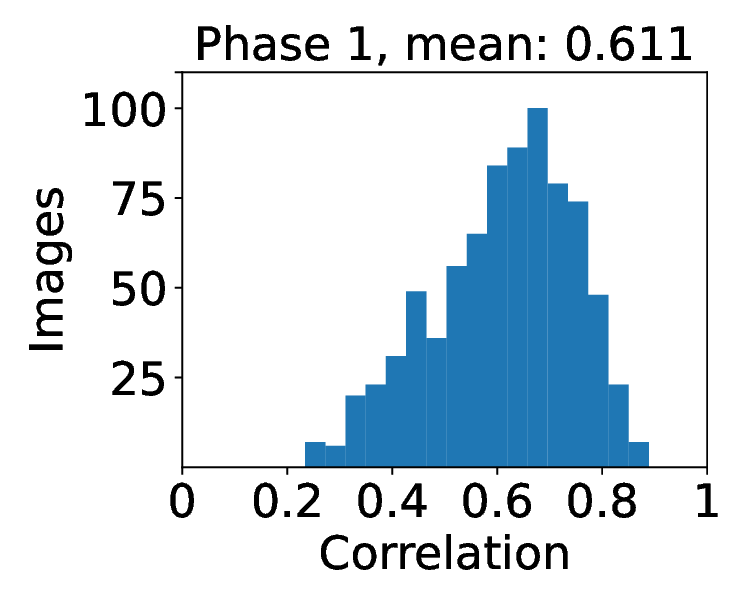}}
\hfil
\subfloat{%
       \includegraphics[scale=0.35]{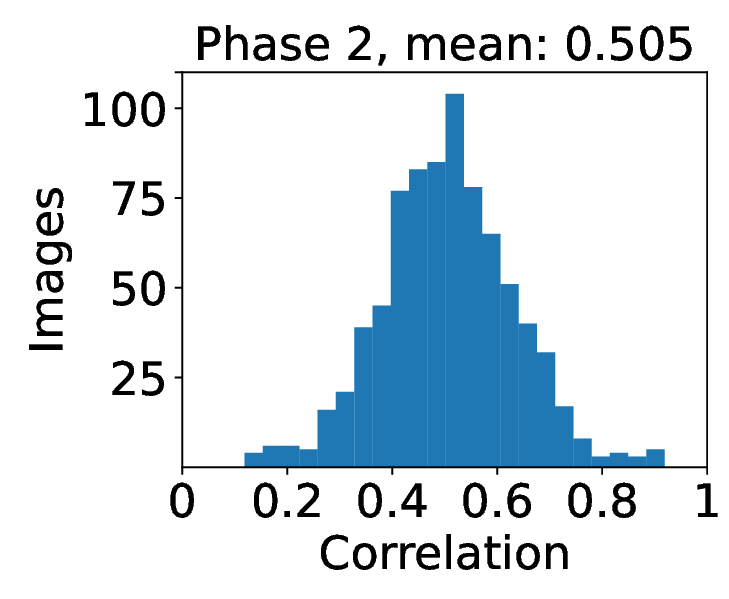}}
\hfil
\subfloat{%
       \includegraphics[scale=0.35]{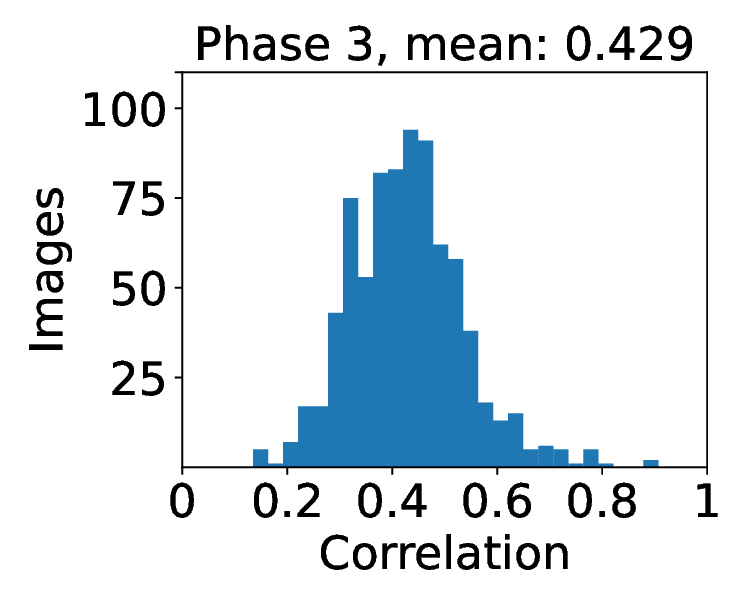}}
\hfil
\subfloat{%
       \includegraphics[scale=0.35]{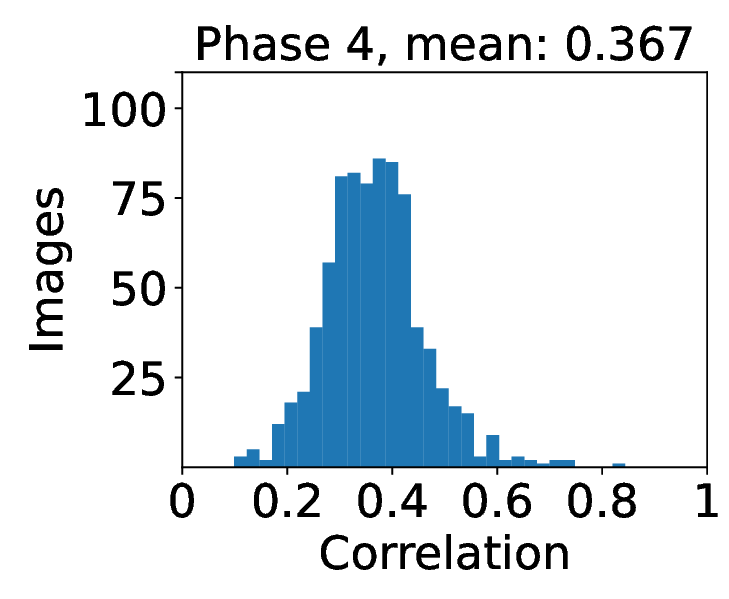}}
\caption{The correlation coefficient between the squared error and the variance of ensemble predictions across phases. For each phase, the correlation is computed for each image in our KITTI test set using predictions by the NN-based construction (RGBd, 1024 samples). The per-phase distributions over images show high correlation that decreases with each phase.}
\label{fig:kitti_1024_correlations}
\end{figure}

\begin{figure}[htbp]
\centering
\includegraphics[width=3.49in]{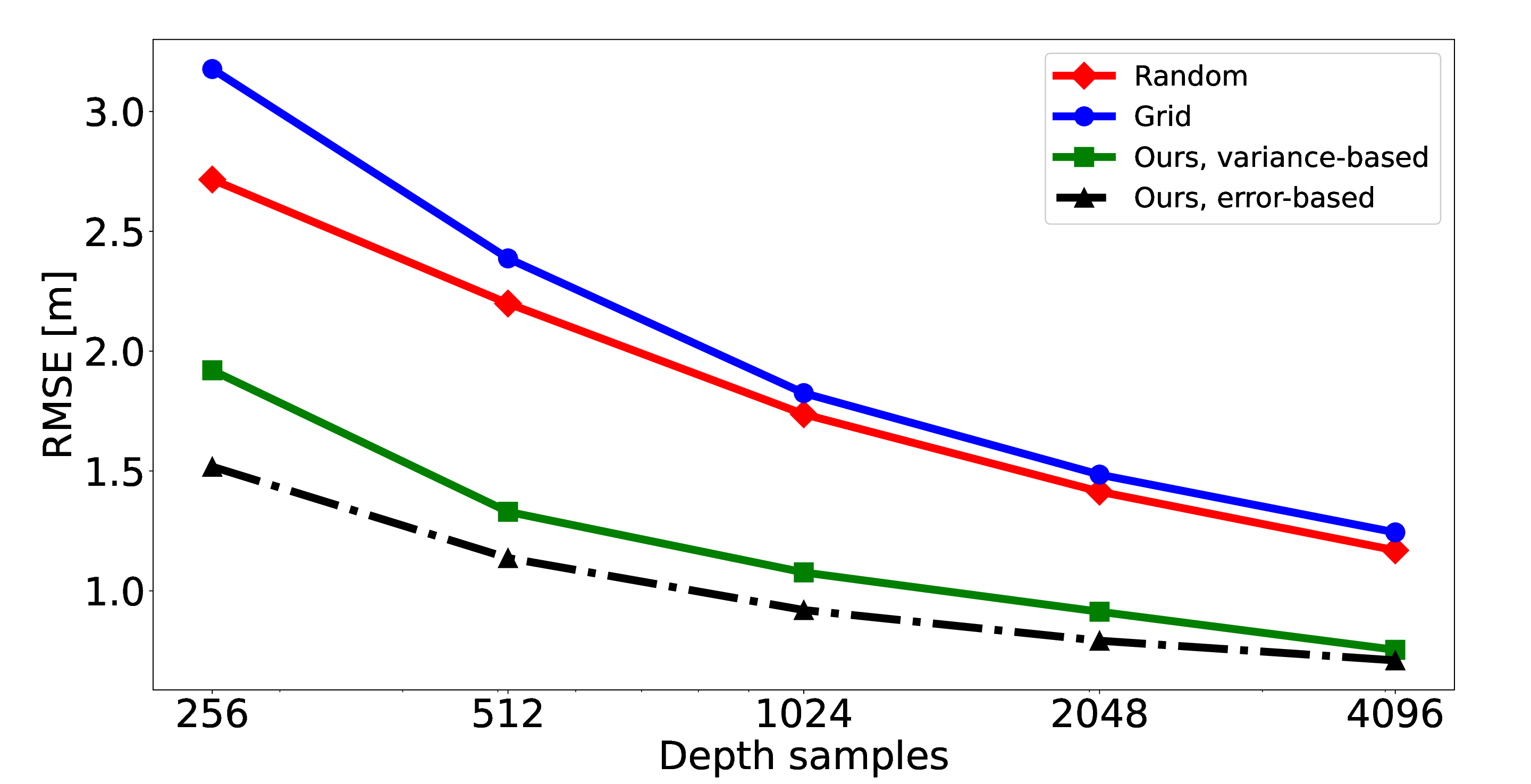}
\caption{An idealized sampling procedure based on prediction squared error instead of variance, compared with the variance-based PM as well as random and grid sampling. Results shown are for our NN-based construction on the KITTI subset with RGB and depth data.} 
\label{fig:kitti_metrics_sparsity_d}
\end{figure}

\begin{figure}[htbp]
\centering
\includegraphics[width=3.46in
]{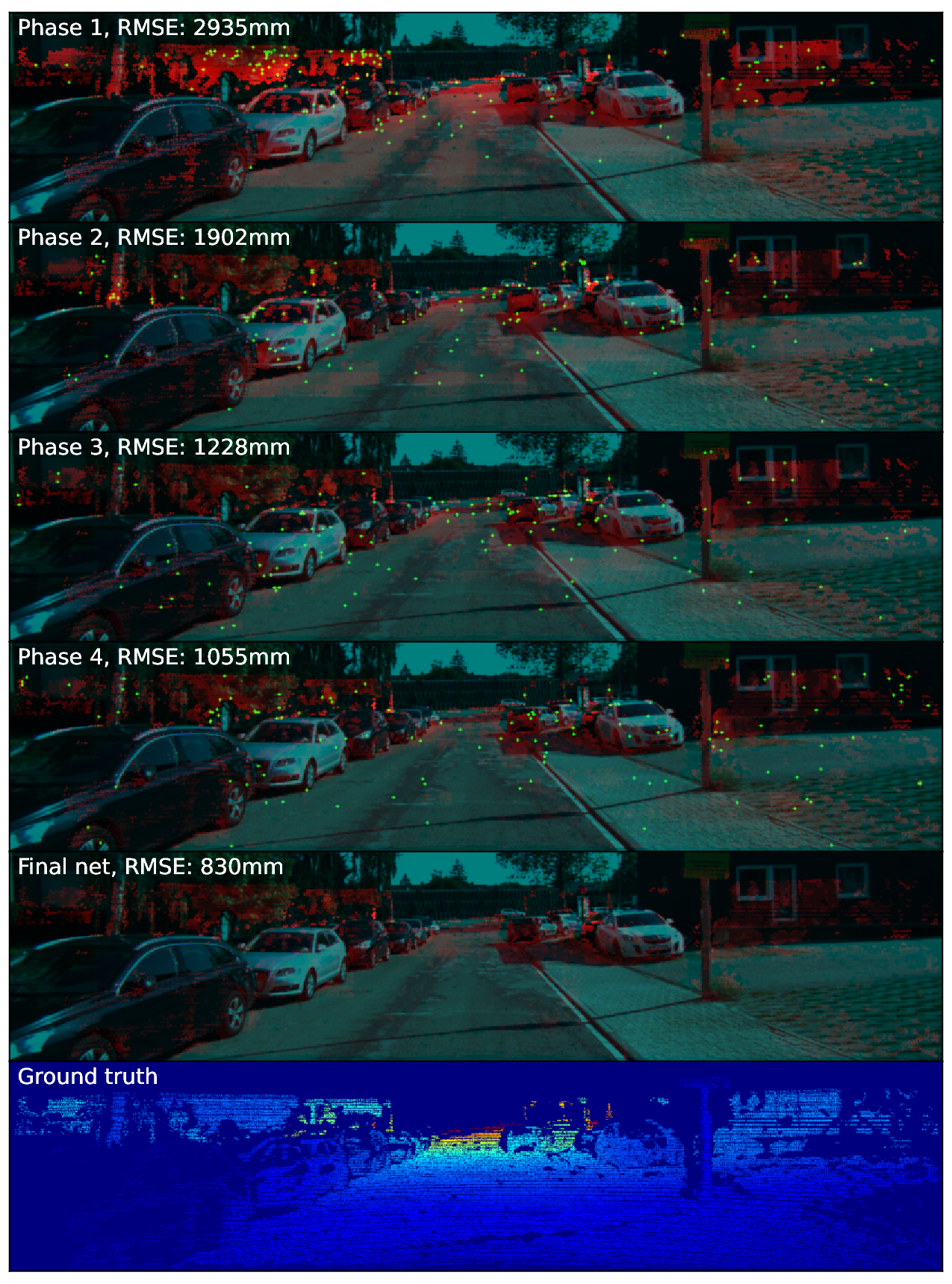}
\caption{Variance-based sampling patterns (green) and error maps (red) across phases for our NN-based construction with RGBd data and 512 samples. Areas of high error are sampled more frequently, reflecting a correlation between the error and the variance and showing a gradual reduction in error. The last error map shows the error of the final predictor.} 
\label{fig:error_during_stages}
\end{figure}

Finally, we examine the relations between the variance and squared error of ensemble predictions in an empirical way. As seen in Figure~\ref{fig:kitti_1024_correlations}, these two quantities are highly correlated, and the correlation decreases with every phase. Namely, choosing points to measure according to their variance is an approximation for choosing them according to their squared error. While the squared error is not available to a sampling algorithm in practice, one may simulate replacing the variance with the squared error in our algorithm, as shown in Figure~\ref{fig:kitti_metrics_sparsity_d}. Interestingly, the RMSE of the idealized method behaves as a lower bound to that of its realistic approximation, the PM method. This behavior, however, is not claimed to always hold. Figure~\ref{fig:error_during_stages} summarizes the relations between error and variance in a visual way by providing a combined view of the variance-based PM sampling pattern and the error across phases.

\subsection{Ablation Studies}\label{subsec:Ablation Studies}
This subsection describes several elements of algorithmic tuning involved in the development of the methods used.

\textbf{The number of phases.} In principle, the number of phases may be as high as the sampling budget, but each additional phase incurs further computational cost, and too many phases may also cause overfitting. Increasing the number of phases improves accuracy dramatically for a small number of phases, but improvement then tapers off or stops. Table~\ref{tab:ablation_phases} shows the performance of our methods as a function of the number of phases (powers of 2). For the NN method the optimal choice is 4, and increasing the number of phases to 8 harms performance. For RF improvement continues even as the number of phases reaches 32, but slows down. We used 8 phases for RF as a trade-off between accuracy and run-time.
\begin{table*}[htbp]
\caption{The effect of the number of phases on the accuracy of our method (RGBd, 1024 samples)}
\label{tab:ablation_phases}
\centering
\scalebox{1}{
    \begin{tabular}{|c|c|c|c|c|c|c|c|c|}
    \hline
    \multirow[c]{2}{*}{\# Phases} & 
    \multicolumn{2}{c|}{RMSE [mm]}&
    \multicolumn{2}{c|}{MAE [mm]}&
    \multicolumn{2}{c|}{REL}&
    \multicolumn{2}{c|}{$\delta_1$ [$\%$]}\\
    \cline{2-9}
    & NN & RF & NN & RF & NN & RF &NN & RF \\
 
    \hline
    1 & 1551 & 1670 & 669 & 638 & 0.045 & 0.044 & 97.85 & 97.56\\
    2 & 1185 & 1342 & 523 & 495 & 0.034 & 0.033 & 98.95 & 98.68\\
    4 & 1077 & 1151 & 473 & 437 & 0.030 & 0.028 & 99.24 & 99.13\\
    8 & 1165 & 1092 & 574 & 414 & 0.035 & 0.027 & 99.18 & 99.27\\
   16 & --   & 1055 & --  & 399 & --    & 0.025 & --    & 99.35\\
   32 & --   & 1033 & --  & 395 & --    & 0.025 & --    & 99.36\\
    
    \hline
    \end{tabular}}
\end{table*}

\textbf{Recalculating the variance within each phase.} Our implementation calculates the variance of ensemble predictions at the beginning of each phase, and uses the result to pick the subset of samples allocated for that phase. A more refined alternative might be to divide the phase into multiple equal sub-phases and recalculate the variance at the beginning of each. This calculation would use predictions based on all samples up to that point, including those in the current phase, without requiring any further retraining. Experiments with the NN-based method revealed no real benefit in using sub-phases, which also incurs a heavier computational cost due to the extra predictions.

\textbf{Settings for the RF-based algorithm.} The size of the final forest, 500, is standard, and large enough for the average of tree predictions to be stable. A similar choice could have been made for all the other forests, but the smaller size of 40 was found to provide sufficient accuracy while allowing for much better run-time. The choice of 3 nearest neighbors in $L_1$, a subsample of 2048 pixels per image, and a few other feature-engineering decisions were made originally using limited experimentation on a set of several hundred images from the Virtual KITTI 1.3.1 dataset \cite{VKITTIGaidonCVPR2016}. Visual inspection ruled out an overlap between those images and the KITTI subset used here.

\section{Conclusion}\label{sec:Conclusion}
In this work we propose a method for adaptive sampling of LiDAR measurements. This becomes an imperative issue as programmable LiDARs, based on solid-state technologies, are introduced into the market. Our solution is based on obtaining the variance of an ensemble of predictors and using it as a proxy for the local error estimation. The sampling stems from a probability density function that is proportional to the ensemble variance (\emph{probability matching}). 
Several sampling phases are introduced to take advantage of partial sampling data and to refine the probability estimations.

We show that the proposed method is significantly superior to grid and to random sampling. Moreover, it outperforms recent state-of-the-art adaptive sampling methods, suggested this year by Bergman et al. \cite{bergman2020deep} and Wolff et al. \cite{wolff2020icra}. The sampling principle can be leveraged by any depth completion algorithm, for which an ensemble can be generated (usually based on learning). Thus it can be used with depth completion algorithms based on neural nets. We also apply it in conjunction with a new, simple depth completion algorithm based on Random Forest. In this setting, the ensemble is immediately available by treating as predictor each tree in the forest. Surprisingly, with adaptive sampling and a fraction of the training set, this Random Forest predictor achieves remarkable results and can be considered as a significant new contribution by itself. 

Future work aims at extending this paradigm to video depth completion, to optimize it to different loss functions, and to establish theoretical guarantees.


%



\section*{Acknowledgment}
We acknowledge support by the Israel Science Foundation (Grant No. 534/19), the Ministry of Science and Technology (Grant 3-15621) and by the Ollendorff Minerva Center.
%

\ifCLASSOPTIONcaptionsoff
  \newpage
\fi

\begin{IEEEbiographynophoto}{Eyal Gofer}
received his Ph.D. from the School of Computer Science, Tel Aviv University in 2014, and was a postdoctoral fellow at the Hebrew University, in the field of machine learning.
In addition, he has held multiple industry positions as a machine learning researcher and algorithm developer in areas spanning text analysis, bioinformatics, and prediction in medicine.
He is a visiting fellow at the Electrical Engineering Department, Technion - Israel Institute of Technology, with research interests that include machine learning, computer vision, and image processing.
\end{IEEEbiographynophoto}
\begin{IEEEbiographynophoto}{Shachar Praisler}
received his B.Sc in Electrical Engineering from the Technion - Israel Institute of Technology in 2019. He is currently working toward M.Sc in Electrical Engineering at the Technion. His research interests are in the fields of computer vision, machine learning and image processing.
\end{IEEEbiographynophoto}
\begin{IEEEbiographynophoto}{Guy Gilboa}
received his Ph.D. from the Electrical Engineering Department,
Technion - Israel Institute of Technology in 2004. He was a postdoctoral fellow
with UCLA and had various development and research roles with Microsoft
and Philips Healthcare. Since 2013 he is a faculty member with the Electrical
Engineering Department, Technion - Israel Institute of Technology. He has
authored some highly cited papers on topics such as image sharpening and
denoising, nonlocal operators theory, and texture analysis. He received several
prizes, including the Eshkol Prize by the Israeli Ministry of Science, the Vatat
Scholarship, and the Gutwirth Prize. He has been serving at the editorial boards of the
journals IEEE SPL, JMIV and CVIU.
\end{IEEEbiographynophoto}








\begin{thebibliography}{10}
\providecommand{\url}[1]{#1}
\csname url@samestyle\endcsname
\providecommand{\newblock}{\relax}
\providecommand{\bibinfo}[2]{#2}
\providecommand{\BIBentrySTDinterwordspacing}{\spaceskip=0pt\relax}
\providecommand{\BIBentryALTinterwordstretchfactor}{4}
\providecommand{\BIBentryALTinterwordspacing}{\spaceskip=\fontdimen2\font plus
\BIBentryALTinterwordstretchfactor\fontdimen3\font minus
  \fontdimen4\font\relax}
\providecommand{\BIBforeignlanguage}[2]{{%
\expandafter\ifx\csname l@#1\endcsname\relax
\typeout{** WARNING: IEEEtran.bst: No hyphenation pattern has been}%
\typeout{** loaded for the language `#1'. Using the pattern for}%
\typeout{** the default language instead.}%
\else
\language=\csname l@#1\endcsname
\fi
#2}}
\providecommand{\BIBdecl}{\relax}
\BIBdecl

\bibitem{cheben2018subwavelength}
P.~Cheben, R.~Halir, J.~H. Schmid, H.~A. Atwater, and D.~R. Smith,
  ``Subwavelength integrated photonics,'' \emph{Nature}, vol. 560, no. 7720,
  pp. 565--572, 2018.

\bibitem{poulton2017coherent}
C.~V. Poulton, A.~Yaacobi, D.~B. Cole, M.~J. Byrd, M.~Raval, D.~Vermeulen, and
  M.~R. Watts, ``Coherent solid-state lidar with silicon photonic optical
  phased arrays,'' \emph{Optics letters}, vol.~42, no.~20, pp. 4091--4094,
  2017.

\bibitem{chapelle2011empirical}
O.~Chapelle and L.~Li, ``An empirical evaluation of thompson sampling,'' in
  \emph{Advances in neural information processing systems}, 2011, pp.
  2249--2257.

\bibitem{Uhrig2017THREEDV}
J.~Uhrig, N.~Schneider, L.~Schneider, U.~Franke, T.~Brox, and A.~Geiger,
  ``Sparsity invariant cnns,'' in \emph{International Conference on 3D Vision
  (3DV)}, 2017.

\bibitem{wolff2020icra}
A.~Wolff, S.~Praisler, I.~Tcenov, and G.~Gilboa, ``Super-pixel sampler: a
  data-driven approach for depth sampling and reconstruction,'' in \emph{2020
  IEEE International Conference on Robotics and Automation (ICRA)}.\hskip 1em
  plus 0.5em minus 0.4em\relax IEEE, 2020.

\bibitem{bergman2020deep}
A.~W. Bergman, D.~B. Lindell, and G.~Wetzstein, ``Deep adaptive lidar:
  End-to-end optimization of sampling and depth completion at low sampling
  rates,'' in \emph{2020 IEEE International Conference on Computational
  Photography (ICCP)}.\hskip 1em plus 0.5em minus 0.4em\relax IEEE, 2020, pp.
  1--11.

\bibitem{park2010depth}
I.~Park and H.~Byun, ``Depth map refinement using multiple patch-based depth
  image completion via local stereo warping,'' \emph{Optical Engineering},
  vol.~49, no.~7, p. 077003, 2010.

\bibitem{park2014high}
J.~Park, H.~Kim, Y.-W. Tai, M.~S. Brown, and I.~S. Kweon, ``High-quality depth
  map upsampling and completion for rgb-d cameras,'' \emph{IEEE Transactions on
  Image Processing}, vol.~23, no.~12, pp. 5559--5572, 2014.

\bibitem{lu2015sparse}
J.~Lu and D.~Forsyth, ``Sparse depth super resolution,'' in \emph{Proceedings
  of the IEEE Conference on Computer Vision and Pattern Recognition}, 2015, pp.
  2245--2253.

\bibitem{wang2008stereoscopic}
L.~Wang, H.~Jin, R.~Yang, and M.~Gong, ``Stereoscopic inpainting: Joint color
  and depth completion from stereo images,'' in \emph{2008 IEEE Conference on
  Computer Vision and Pattern Recognition}.\hskip 1em plus 0.5em minus
  0.4em\relax IEEE, 2008, pp. 1--8.

\bibitem{camplani2012efficient}
M.~Camplani and L.~Salgado, ``Efficient spatio-temporal hole filling strategy
  for kinect depth maps,'' in \emph{Three-dimensional image processing (3DIP)
  and applications II}, vol. 8290.\hskip 1em plus 0.5em minus 0.4em\relax
  International Society for Optics and Photonics, 2012, p. 82900E.

\bibitem{shen2013layer}
J.~Shen and S.-C.~S. Cheung, ``Layer depth denoising and completion for
  structured-light rgb-d cameras,'' in \emph{Proceedings of the IEEE conference
  on computer vision and pattern recognition}, 2013, pp. 1187--1194.

\bibitem{lu2014depth}
S.~Lu, X.~Ren, and F.~Liu, ``Depth enhancement via low-rank matrix
  completion,'' in \emph{Proceedings of the IEEE conference on computer vision
  and pattern recognition}, 2014, pp. 3390--3397.

\bibitem{drozdov2016robust}
G.~Drozdov, Y.~Shapiro, and G.~Gilboa, ``Robust recovery of heavily degraded
  depth measurements,'' in \emph{2016 Fourth International Conference on 3D
  Vision (3DV)}.\hskip 1em plus 0.5em minus 0.4em\relax IEEE, 2016, pp. 56--65.

\bibitem{ma2016sparse}
F.~Ma, L.~Carlone, U.~Ayaz, and S.~Karaman, ``Sparse sensing for
  resource-constrained depth reconstruction,'' in \emph{2016 IEEE/RSJ
  International Conference on Intelligent Robots and Systems (IROS)}.\hskip 1em
  plus 0.5em minus 0.4em\relax IEEE, 2016, pp. 96--103.

\bibitem{ku2018defense}
J.~Ku, A.~Harakeh, and S.~L. Waslander, ``In defense of classical image
  processing: Fast depth completion on the cpu,'' in \emph{2018 15th Conference
  on Computer and Robot Vision (CRV)}.\hskip 1em plus 0.5em minus 0.4em\relax
  IEEE, 2018, pp. 16--22.

\bibitem{ma2018sparse}
F.~Ma and S.~Karaman, ``Sparse-to-dense: Depth prediction from sparse depth
  samples and a single image,'' in \emph{2018 IEEE International Conference on
  Robotics and Automation (ICRA)}.\hskip 1em plus 0.5em minus 0.4em\relax IEEE,
  2018, pp. 1--8.

\bibitem{chen2018estimating}
Z.~Chen, V.~Badrinarayanan, G.~Drozdov, and A.~Rabinovich, ``Estimating depth
  from rgb and sparse sensing,'' in \emph{Proceedings of the European
  Conference on Computer Vision (ECCV)}, 2018, pp. 167--182.

\bibitem{cheng2018depth}
X.~Cheng, P.~Wang, and R.~Yang, ``Depth estimation via affinity learned with
  convolutional spatial propagation network,'' in \emph{Proceedings of the
  European Conference on Computer Vision (ECCV)}, 2018, pp. 103--119.

\bibitem{wang2018multi}
B.~Wang, Y.~Feng, and H.~Liu, ``Multi-scale features fusion from sparse lidar
  data and single image for depth completion,'' \emph{Electronics Letters},
  vol.~54, no.~24, pp. 1375--1377, 2018.

\bibitem{shivakumar2019dfusenet}
S.~S. Shivakumar, T.~Nguyen, I.~D. Miller, S.~W. Chen, V.~Kumar, and C.~J.
  Taylor, ``Dfusenet: Deep fusion of rgb and sparse depth information for image
  guided dense depth completion,'' in \emph{2019 IEEE Intelligent
  Transportation Systems Conference (ITSC)}.\hskip 1em plus 0.5em minus
  0.4em\relax IEEE, 2019, pp. 13--20.

\bibitem{huang2019hms}
Z.~Huang, J.~Fan, S.~Cheng, S.~Yi, X.~Wang, and H.~Li, ``Hms-net: Hierarchical
  multi-scale sparsity-invariant network for sparse depth completion,''
  \emph{IEEE Transactions on Image Processing}, 2019.

\bibitem{chen2019learning}
Y.~Chen, B.~Yang, M.~Liang, and R.~Urtasun, ``Learning joint 2d-3d
  representations for depth completion,'' in \emph{Proceedings of the IEEE
  International Conference on Computer Vision}, 2019, pp. 10\,023--10\,032.

\bibitem{tang2019learning}
J.~Tang, F.-P. Tian, W.~Feng, J.~Li, and P.~Tan, ``Learning guided
  convolutional network for depth completion,'' \emph{arXiv preprint
  arXiv:1908.01238}, 2019.

\bibitem{lee2020deep}
S.~Lee, J.~Lee, D.~Kim, and J.~Kim, ``Deep architecture with cross guidance
  between single image and sparse lidar data for depth completion,'' \emph{IEEE
  Access}, vol.~8, pp. 79\,801--79\,810, 2020.

\bibitem{lee2019depth}
B.-U. Lee, H.-G. Jeon, S.~Im, and I.~S. Kweon, ``Depth completion with deep
  geometry and context guidance,'' in \emph{2019 International Conference on
  Robotics and Automation (ICRA)}.\hskip 1em plus 0.5em minus 0.4em\relax IEEE,
  2019, pp. 3281--3287.

\bibitem{eldesokey2019confidence}
A.~Eldesokey, M.~Felsberg, and F.~S. Khan, ``Confidence propagation through
  cnns for guided sparse depth regression,'' \emph{IEEE transactions on pattern
  analysis and machine intelligence}, 2019.

\bibitem{van2019sparse}
W.~Van~Gansbeke, D.~Neven, B.~De~Brabandere, and L.~Van~Gool, ``Sparse and
  noisy lidar completion with rgb guidance and uncertainty,'' in \emph{2019
  16th International Conference on Machine Vision Applications (MVA)}.\hskip
  1em plus 0.5em minus 0.4em\relax IEEE, 2019, pp. 1--6.

\bibitem{qiu2019deeplidar}
J.~Qiu, Z.~Cui, Y.~Zhang, X.~Zhang, S.~Liu, B.~Zeng, and M.~Pollefeys,
  ``Deeplidar: Deep surface normal guided depth prediction for outdoor scene
  from sparse lidar data and single color image,'' in \emph{Proceedings of the
  IEEE Conference on Computer Vision and Pattern Recognition}, 2019, pp.
  3313--3322.

\bibitem{xu2019depth}
Y.~Xu, X.~Zhu, J.~Shi, G.~Zhang, H.~Bao, and H.~Li, ``Depth completion from
  sparse lidar data with depth-normal constraints,'' in \emph{Proceedings of
  the IEEE International Conference on Computer Vision}, 2019, pp. 2811--2820.

\bibitem{ma2019self}
F.~Ma, G.~V. Cavalheiro, and S.~Karaman, ``Self-supervised sparse-to-dense:
  Self-supervised depth completion from lidar and monocular camera,'' in
  \emph{2019 International Conference on Robotics and Automation (ICRA)}.\hskip
  1em plus 0.5em minus 0.4em\relax IEEE, 2019, pp. 3288--3295.

\bibitem{qu2020depth}
C.~Qu, T.~Nguyen, and C.~Taylor, ``Depth completion via deep basis fitting,''
  in \emph{The IEEE Winter Conference on Applications of Computer Vision},
  2020, pp. 71--80.

\bibitem{eldar1997farthest}
Y.~Eldar, M.~Lindenbaum, M.~Porat, and Y.~Y. Zeevi, ``The farthest point
  strategy for progressive image sampling,'' \emph{IEEE Transactions on Image
  Processing}, vol.~6, no.~9, pp. 1305--1315, 1997.

\bibitem{zhu2015adaptive}
S.~Zhu, B.~Zeng, and M.~Gabbouj, ``Adaptive sampling for compressed sensing
  based image compression,'' \emph{Journal of Visual Communication and Image
  Representation}, vol.~30, pp. 94--105, 2015.

\bibitem{dovrat2019learning}
O.~Dovrat, I.~Lang, and S.~Avidan, ``Learning to sample,'' in \emph{Proceedings
  of the IEEE Conference on Computer Vision and Pattern Recognition}, 2019, pp.
  2760--2769.

\bibitem{dai2019adaptive}
Q.~Dai, H.~Chopp, E.~Pouyet, O.~Cossairt, M.~Walton, and A.~Katsaggelos,
  ``Adaptive image sampling using deep learning and its application on x-ray
  fluorescence image reconstruction,'' \emph{IEEE Transactions on Multimedia},
  2019.

\bibitem{hawe2011dense}
S.~Hawe, M.~Kleinsteuber, and K.~Diepold, ``Dense disparity maps from sparse
  disparity measurements,'' in \emph{2011 International Conference on Computer
  Vision}.\hskip 1em plus 0.5em minus 0.4em\relax IEEE, 2011, pp. 2126--2133.

\bibitem{liu2015depth}
L.-K. Liu, S.~H. Chan, and T.~Q. Nguyen, ``Depth reconstruction from sparse
  samples: Representation, algorithm, and sampling,'' \emph{IEEE Transactions
  on Image Processing}, vol.~24, no.~6, pp. 1983--1996, 2015.

\bibitem{sun2013large}
J.~Sun, E.~Timurdogan, A.~Yaacobi, E.~S. Hosseini, and M.~R. Watts,
  ``Large-scale nanophotonic phased array,'' \emph{Nature}, vol. 493, no. 7431,
  pp. 195--199, 2013.

\bibitem{tasneem2018directionally}
Z.~Tasneem, D.~Wang, H.~Xie, and K.~Sanjeev, ``Directionally controlled
  time-of-flight ranging for mobile sensing platforms.'' in \emph{Robotics:
  Science and Systems}, 2018.

\bibitem{hannekeActiveSurvey}
S.~Hanneke, ``Theory of disagreement-based active learning,'' \emph{Foundations
  and Trends® in Machine Learning}, vol.~7, no. 2-3, pp. 131--309, 2014.

\bibitem{settles2009active}
B.~Settles, ``Active learning literature survey,'' University of
  Wisconsin-Madison Department of Computer Sciences, Madison, WI, Tech. Rep.
  1648, Jan. 2009.

\bibitem{liu2018survey}
H.~Liu, Y.-S. Ong, and J.~Cai, ``A survey of adaptive sampling for global
  metamodeling in support of simulation-based complex engineering design,''
  \emph{Structural and Multidisciplinary Optimization}, vol.~57, no.~1, pp.
  393--416, 2018.

\bibitem{vandoni2019evidential}
J.~Vandoni, E.~Aldea, and S.~Le~H{\'e}garat-Mascle, ``Evidential
  query-by-committee active learning for pedestrian detection in high-density
  crowds,'' \emph{International Journal of Approximate Reasoning}, vol. 104,
  pp. 166--184, 2019.

\bibitem{polewski2015active}
P.~Polewski, W.~Yao, M.~Heurich, P.~Krzystek, and U.~Stilla, ``Active learning
  approach to detecting standing dead trees from als point clouds combined with
  aerial infrared imagery,'' in \emph{Proceedings of the IEEE Conference on
  Computer Vision and Pattern Recognition Workshops}, 2015, pp. 10--18.

\bibitem{konyushkova2019geometry}
K.~Konyushkova, R.~Sznitman, and P.~Fua, ``Geometry in active learning for
  binary and multi-class image segmentation,'' \emph{Computer Vision and Image
  Understanding}, 2019.

\bibitem{gu2017embedded}
Q.~Gu, J.~Yang, L.~Kong, W.~Q. Yan, and R.~Klette, ``Embedded and real-time
  vehicle detection system for challenging on-road scenes,'' \emph{Optical
  Engineering}, vol.~56, no.~6, p. 063102, 2017.

\bibitem{jiang2018integrating}
T.~Jiang, Y.~Wang, S.~Tao, Y.~Li, and S.~Liu, ``Integrating active learning and
  contextually guide for semantic labeling of lidar point cloud,'' in
  \emph{2018 10th IAPR Workshop on Pattern Recognition in Remote Sensing
  (PRRS)}.\hskip 1em plus 0.5em minus 0.4em\relax IEEE, 2018, pp. 1--7.

\bibitem{feng2019deep}
D.~Feng, X.~Wei, L.~Rosenbaum, A.~Maki, and K.~Dietmayer, ``Deep active
  learning for efficient training of a lidar 3d object detector,'' \emph{arXiv
  preprint arXiv:1901.10609}, 2019.

\bibitem{rai2019monocular}
A.~Rai, ``Monocular depth estimation with edge-based constraints and active
  learning,'' Master's thesis, Arizona State University, 2019.

\bibitem{cohn1996active}
D.~A. Cohn, Z.~Ghahramani, and M.~I. Jordan, ``Active learning with statistical
  models,'' \emph{Journal of artificial intelligence research}, vol.~4, pp.
  129--145, 1996.

\bibitem{cohn1997minimizing}
D.~A. Cohn, ``Minimizing statistical bias with queries,'' in \emph{Advances in
  neural information processing systems}, 1997, pp. 417--423.

\bibitem{seung1992query}
H.~S. Seung, M.~Opper, and H.~Sompolinsky, ``Query by committee,'' in
  \emph{Proceedings of the fifth annual workshop on Computational learning
  theory}.\hskip 1em plus 0.5em minus 0.4em\relax ACM, 1992, pp. 287--294.

\bibitem{freund1997selective}
Y.~Freund, H.~S. Seung, E.~Shamir, and N.~Tishby, ``Selective sampling using
  the query by committee algorithm,'' \emph{Machine learning}, vol.~28, no.
  2-3, pp. 133--168, 1997.

\bibitem{AbeM98}
N.~Abe and H.~Mamitsuka, ``Query learning strategies using boosting and
  bagging,'' in \emph{Proceedings of the Fifteenth International Conference on
  Machine Learning ({ICML})}.\hskip 1em plus 0.5em minus 0.4em\relax Morgan
  Kaufmann, 1998, pp. 1--9.

\bibitem{burbidge2007active}
R.~Burbidge, J.~J. Rowland, and R.~D. King, ``Active learning for regression
  based on query by committee,'' in \emph{International Conference on
  Intelligent Data Engineering and Automated Learning}.\hskip 1em plus 0.5em
  minus 0.4em\relax Springer, 2007, pp. 209--218.

\bibitem{borisov2011active}
A.~Borisov, E.~Tuv, and G.~Runger, ``Active batch learning with stochastic
  query-by-forest (sqbf),'' in \emph{Active Learning and Experimental Design
  workshop In conjunction with AISTATS 2010}, 2011, pp. 59--69.

\bibitem{douak2012active}
F.~Douak, F.~Melgani, N.~Alajlan, E.~Pasolli, Y.~Bazi, and N.~Benoudjit,
  ``Active learning for spectroscopic data regression,'' \emph{Journal of
  Chemometrics}, vol.~26, no.~7, pp. 374--383, 2012.

\bibitem{kee2018query}
S.~Kee, E.~del Castillo, and G.~Runger, ``Query-by-committee improvement with
  diversity and density in batch active learning,'' \emph{Information
  Sciences}, vol. 454, pp. 401--418, 2018.

\bibitem{wu2018pool}
D.~Wu, ``Pool-based sequential active learning for regression,'' \emph{IEEE
  transactions on neural networks and learning systems}, 2018.

\bibitem{park2019active}
S.~H. Park and S.~B. Kim, ``Active semi-supervised learning with multiple
  complementary information,'' \emph{Expert Systems with Applications}, vol.
  126, pp. 30--40, 2019.

\bibitem{roy2001toward}
N.~Roy and A.~McCallum, ``Toward optimal active learning through monte carlo
  estimation of error reduction,'' \emph{ICML, Williamstown}, pp. 441--448,
  2001.

\bibitem{konyushkova2017learning}
K.~Konyushkova, R.~Sznitman, and P.~Fua, ``Learning active learning from
  data,'' in \emph{Advances in Neural Information Processing Systems}, 2017,
  pp. 4225--4235.

\bibitem{kading2018active}
C.~K{\"a}ding, E.~Rodner, A.~Freytag, O.~Mothes, B.~Barz, J.~Denzler, and C.~Z.
  AG, ``Active learning for regression tasks with expected model output
  changes,'' in \emph{British Machine Vision Conference (BMVC)}, 2018.

\bibitem{aute2013cross}
V.~Aute, K.~Saleh, O.~Abdelaziz, S.~Azarm, and R.~Radermacher,
  ``Cross-validation based single response adaptive design of experiments for
  kriging metamodeling of deterministic computer simulations,''
  \emph{Structural and Multidisciplinary Optimization}, vol.~48, no.~3, pp.
  581--605, 2013.

\bibitem{willett2006faster}
R.~Willett, R.~Nowak, and R.~M. Castro, ``Faster rates in regression via active
  learning,'' in \emph{Advances in Neural Information Processing Systems},
  2006, pp. 179--186.

\bibitem{goetz2018active}
J.~Goetz, A.~Tewari, and P.~Zimmerman, ``Active learning for non-parametric
  regression using purely random trees,'' in \emph{Advances in Neural
  Information Processing Systems}, 2018, pp. 2537--2546.

\bibitem{Breiman2001}
L.~Breiman, ``Random forests,'' \emph{Machine Learning}, vol.~45, no.~1, pp.
  5--32, Oct 2001.

\bibitem{scikit-learn}
F.~Pedregosa, G.~Varoquaux, A.~Gramfort, V.~Michel, B.~Thirion, O.~Grisel,
  M.~Blondel, P.~Prettenhofer, R.~Weiss, V.~Dubourg, J.~Vanderplas, A.~Passos,
  D.~Cournapeau, M.~Brucher, M.~Perrot, and E.~Duchesnay, ``Scikit-learn:
  Machine learning in {P}ython,'' \emph{Journal of Machine Learning Research},
  vol.~12, pp. 2825--2830, 2011.

\bibitem{VKITTIGaidonCVPR2016}
A.~Gaidon, Q.~Wang, Y.~Cabon, and E.~Vig, ``Virtual worlds as proxy for
  multi-object tracking analysis,'' in \emph{CVPR}, 2016.

\end{thebibliography}
\end{document}